\documentclass[runningheads]{llncs}

\usepackage{eccv}

\usepackage{eccvabbrv}

\usepackage{graphicx}
\usepackage{booktabs}
\usepackage{soul}
\usepackage{placeins}

\usepackage[accsupp]{axessibility}  % Improves PDF readability for those with disabilities.

\usepackage{hyperref}

\usepackage{xr-hyper}
\begin{document}

% ---------------------------------------------------------------
\title{Find the Assembly Mistakes: \\Error Segmentation for Industrial Applications} 

\titlerunning{Find the Assembly Mistakes: Error Segmentation for Industrial Applications}

\author{Dan Lehman$^\textbf{*}$\inst{1,2} \and
Tim J. Schoonbeek$^\textbf{*}$\inst{1,2} \and
Shao-Hsuan Hung\inst{1,2} \and
Jacek Kustra\inst{2} \and \\
Peter H.N. de With\inst{1} \and
Fons van der Sommen\inst{1} \\
\vspace{0.1cm}
{\small $^\textbf{*}$Equal contribution (ordered by coin-flip)}
}

\authorrunning{D.~Lehman and T.J.~Schoonbeek et al.}

\institute{Eindhoven University of Technology, Eindhoven, The Netherlands \and
ASML Research, Veldhoven, The Netherlands}

\maketitle

\begin{abstract}
  Recognizing errors in assembly and maintenance procedures is valuable for industrial applications, since it can increase worker efficiency and prevent unplanned down-time. Although assembly state recognition is gaining attention, none of the current works investigate assembly error \emph{localization}. Therefore, we propose StateDiffNet, which localizes assembly errors based on detecting the differences between a~(correct) intended assembly state and a test image from a similar viewpoint. StateDiffNet is trained on synthetically generated image pairs, providing full control over the type of meaningful change that should be detected. The proposed approach is the first to correctly localize assembly errors taken from real ego-centric video data for both states and error types that are never presented during training. Furthermore, the deployment of change detection to this industrial application provides valuable insights and considerations into the mechanisms of state-of-the-art change detection algorithms. The code and data generation pipeline are publicly available at: 
   {\url{https://timschoonbeek.github.io/error_seg}}.
\end{abstract}

\section{Introduction}
\label{sec:intro}

Manual assembly and maintenance tasks are an important part of high-tech systems and an integral part of the job of a service engineer. In many industrial settings, (dis)assembly errors can have costly consequences, such as extended down-time.
To avoid errors, workers follow paper manuals to interpret work instructions, but this is time-consuming and imposes a cognitive burden on the operator~\cite{li-2024}. Besides, it is not scalable to the assembly of increasingly complex systems. Meanwhile, recent advancements in augmented reality~(AR) technologies are providing the potential to replace paper manuals and provide real-time assistance during assembly and maintenance tasks. This development has sparked interest in researching the use of sensor data from AR headsets to automatically verify and log procedure steps, to recognize potential errors~\cite{industreal, li-2024, prego, asdf, albers2023augmented}.

\begin{figure}
    \centering
    \includegraphics[width=0.99\linewidth]{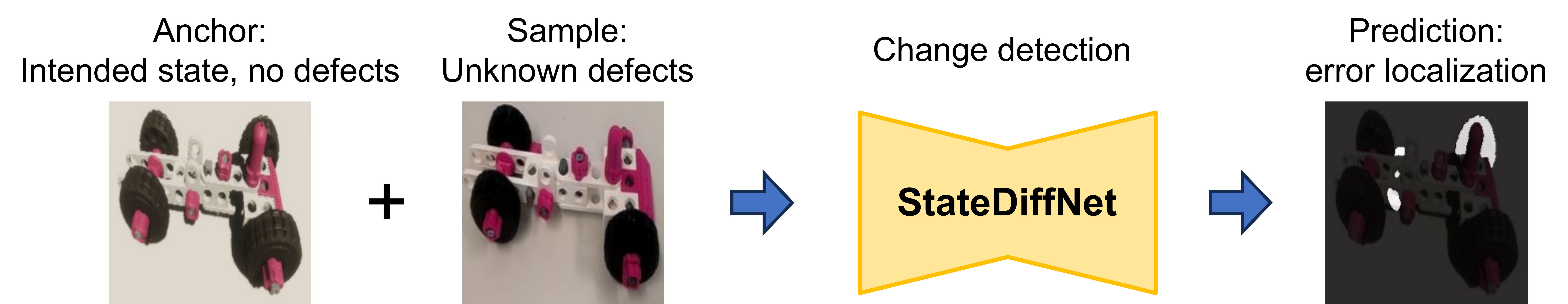}
    \caption{Principal view of the proposed method on an object from a complex assembly procedure~\cite{industreal}. The approach is evaluated on unseen assembly states and error types.}
    \label{fig:teaser}
    \vspace{-0.5cm}
\end{figure}

Many works perform some type of assembly state detection~\cite{industreal, asdf, li-2024, coffeeMachine, state_aware_detection} or assembly quality inspection~\cite{albers2023augmented}, which are typically performed as a supervised classification or detection problem. Though these approaches show decent performance at classifying a small number of assembly states, performance on erroneous states is not investigated. A representation-learning based approach, rather than a classification based approach, is demonstrated to be capable of recognizing \textit{unseen} assembly errors~\cite{simstate}, a fundamental requirement for a viable system. However, this approach lacks spatial localization of errors, and therefore cannot provide interpretable feedback. Another type of approach relies on anomaly detection, which requires a model to learn the typical structures of nominal data, based on a-priori definitions of nominal data~\cite{reconpatch, vision-datasets}. This is not feasible for industrial procedures, since the possible presence of an error in assembly states depends on the intended state at that moment of the assembly. Accordingly, change detection has been proposed to segment progress between various assembly states~\cite{AssCD_real, AssCD_synth}. However, their method is applied only to non-erroneous assemblies and to a highly simplified task, with only four procedure steps in the entire assembly.

To overcome the fundamental limitations that arise for error localization in the industrial domain, we propose a methodology, named StateDiffNet, that pinpoints the difference between an object in different assembly~(including erroneous) states, using segmentation.
StateDiffNet, outlined in~\autoref{fig:teaser}, is the first error localization system that can locate errors on states that were never encountered during training, and on much more complex assembly configurations than related works. For example, where the closest related work trains and evaluates on the same five assembly states~\cite{AssCD_real}, we use more than 10$^5$ unique state combinations, including states with very small differences, and test on entirely unseen states. Given two images of an assembly object, StateDiffNet segments all meaningful differences in the state that can be inferred from its view of the object. A core component of the proposed approach is the methodology for generating and sampling synthetic image pairs, which provides full control over the meaningful change that the system should detect, as well as the changes that the system should be invariant to, \ie the expected variability resulting from aspects such as camera pose, photometry, image distortions, and shadows. 

We demonstrate that StateDiffNet, trained exclusively on synthetic data, is able to segment assembly errors taken from real ego-centric video data~\cite{industreal}, including various error types that were never presented during training. The proposed method can be adapted to many industrial assembly procedures, provided that a CAD model of the assembly object is available. In summary, the main innovations of this paper consist of the following contributions.
\begin{itemize}
    \item The first assembly error localization network used to study error localization on assemblies with a wide variety of complex state differences. The code is made publicly available.
    \item A data generation technique that, given a 3D~model, creates image pairs for change detection, allowing full control of the expected~(normal) variability between image pairs. The generation pipeline is made publicly available.
    \item Valuable insights into the mechanisms of state-of-the-art change detection algorithms applied specifically to industrial applications.
\end{itemize}

\section{Related Work}
\label{sec:related}
There are multiple existing research fields that investigate tasks similar to assembly error localization, including assembly state recognition and change detection. Relevant works from each of those fields are outlined below.

\subsubsection{Assembly state recognition:}
Assembly~(or object) state detection~(ASD) is a specialized version of object detection, wherein the aim is to identify a bounding box around an object, which is labeled according to its state in the assembly process. Su~\etal~\cite{coffeeMachine} and Schieber~\etal~\cite{asdf} simultaneously perform ASD and 6D~pose estimation, which is particularly useful for AR applications. However, they use assembly procedures of limited complexity~(at most~10 parts, which can only be assembled in a limited number of ways). Two recent works focus on assembly procedures with more part complexity using object detection networks, trained on both synthetic and real-world annotated images~\cite{industreal, state_aware_detection}. However, it is unclear whether the method scales to procedures of higher complexity, containing potentially millions~(as opposed to tens) of different states. Additionally, little effort has been undertaken to investigate how these models handle erroneous assembly states.

To address these limitations, a recent work~\cite{simstate} proposed a representation learning approach towards assembly state recognition. They show that such a model is able to retrieve states that are never encountered during training and classify some states as erroneous, while being trained purely on error-free data. However, the model does not provide any indication on the source of the error, thus providing limited assistance to the operator of such a system. Additionally, this prohibits the~(automatic) recognition of correctly assembled parts if an error is present anywhere on the assembly. The proposed work in this paper aims to solve these limitations, while maintaining the valuable capability to generalize to assembly states entirely unseen during training.

\subsubsection{Change detection:}
The objective of change detection is to detect areas that have incurred \textit{meaningful} change between two images of the same scene taken at two different time instances, and potentially from different viewpoints. This commonly takes the form of binary change segmentation focused on remote sensing~\cite{BIT,TinyCD,ChangeFormer} and surveillance applications~\cite{IED, jst2015change, spot-the-diff_dataset}. The best-performing methods for change detection are deep learning based~\cite{cd_survey}, \eg the Bi-temporal Image Transformer~(BIT)~\cite{BIT}. In remote sensing applications, the two input images to the change detection network are generally taken from the same position using georeferencing and feature matching~\cite{IED}, simplifying a change detection model's task. To address this simplification, Sachdeva and Zisserman~\cite{cyws, cyws3d} introduce a network architecture suited to detect a much more general form of change based on `object-level' difference~(irrespective of the object class), by predicting bounding boxes on both images to indicate change. Their model uses DINO~\cite{DINO} as a Siamese feature extractor and relies on re-projecting image features to a common coordinate system, while ignoring occluded pixels from the novel view. 

Two works apply change detection to manual assemblies~\cite{AssCD_synth,AssCD_real}, demonstrating that the BIT architecture can be used to segment change on assembly objects even under rotations. However, both apply change detection to a simplified dataset, containing images gathered from a fixed top-view camera, and a nearly constant roll, pitch, and scale for the object states. Most importantly, they do not apply change detection for the segmentation of errors, but to distinguish the differences between five highly distinctive, pre-defined object states. Therefore, their model is not capable of generalizing to unseen states, \eg with assembly errors. These constraints greatly simplify their task, making it unsuitable for assembly error localization in many real industrial use cases. \\

\noindent The work in this paper addresses the following limitations of related works:
\vspace{-0.2cm}
\begin{itemize}
    \item \textit{Error localization.} The use change detection for error localization, requiring an entirely different training strategy.
    \item \textit{Object complexity.} The proposed approach is evaluated on the recently introduced IndustReal dataset~\cite{industreal}, which is significantly more complex than prior works, due to the large number of parts~(36) and the high degree of visual similarity between states.
    \item \textit{Generalization.} The proposed work is tested for generalization to assembly states not seen during training as well as entirely unseen error types.
\end{itemize}

\section{Proposed Method}
\label{sec:methods}
Given two images, an anchor and a sample image, containing objects in similar orientations during an assembly or maintenance procedure, our objective is to segment the meaningful change between these two images. This meaningful change corresponds to error segmentation when provided with a known error-free anchor image and a sample image with unknown presence of defects. This segmentation can provide valuable feedback to workers during the execution of assembly procedures. We propose an approach for error segmentation based on a modified change detection algorithm, and a novel approach for image-pair generation that provides full control over the type and degree of change that is considered meaningful.

\begin{figure*}[tb]
    \centering
    \includegraphics[width=0.8\linewidth]{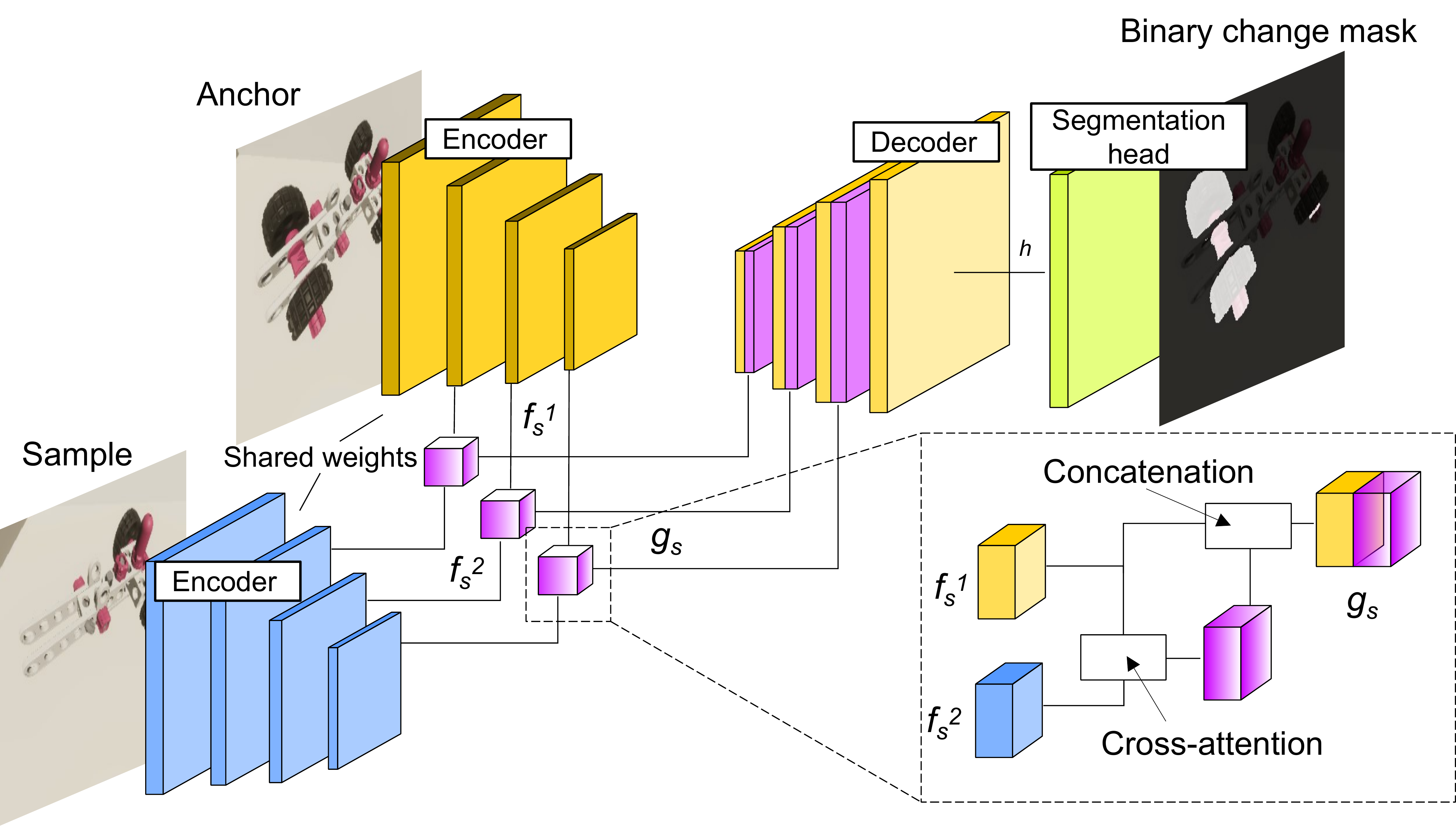}
    \caption{Proposed change segmentation architecture, modified from~\cite{cyws}, consisting of a Siamese encoder, cross-attention based feature-fusion blocks as skip connections in the U-Net-style decoder, and a two-layer convolutional segmentation head. }
    \label{fig:cyws}
    % \vspace{-0.5cm}
\end{figure*}

\subsection{Model Architecture} 
\label{sec:statediffnet}
The model architecture is modified from a state-of-the-art change detection algorithm~\cite{cyws}, consisting of a U-Net-style Siamese encoder and decoder~\cite{unet}. The original bounding-box head is replaced by a segmentation head. Furthermore, one of the decoders is removed, to make predictions on only the anchor image instead of both. The architecture of the model is depicted in \autoref{fig:cyws}. Both the anchor and sample images are independently passed through the same convolutional encoder to obtain feature maps~$f_s^1$ and $f_s^2$ at multiple spatial resolutions~$s$. The decoder is tasked with upsampling and segmentation of the differences in features between~$f_s^1$ and $f_s^2$, to produce a feature map~$h$ at the original resolution of the sample image. For this purpose, each skip connection contains the feature map~$f_s^1$ concatenated with the cross-attention between $f_s^1$ and $f_s^2$, to propagate the features from the sample image to the decoder. 

The use of this cross-attention module, called global cross-attention~(GCA), is the architecture's key component for change detection~\cite{cyws}. For each feature vector~$f_s^1(x,y)$ at spatial location~$(x,y)$, the concatenated output of the cross-attention mechanism contains the sum of all feature vectors in~$f_s^2$, weighted by their similarity to the feature vector $f_s^1(x,y)$. Intuitively, the most similar features of~$f_s^2$ become concatenated to~$f_s^1$ at the same spatial location, allowing the model to predict change by identifying at which spatial locations no matching feature was found in~$f_s^2$, since then the output of the cross-attention is simply some random weighted average of the features in~$f_s^2$. However, there are some inherent fundamental flaws in this cross-attention module, some of which become particularly relevant when employed in the industrial domain. The following paragraphs discuss these flaws and two proposed ways to address them.

\begin{figure*}[tb]
  \centering
  \subfloat[Cross-attention with multi-headed self-attention.\label{fig:MSA+CA}]{%
       \includegraphics[width=0.6\linewidth]{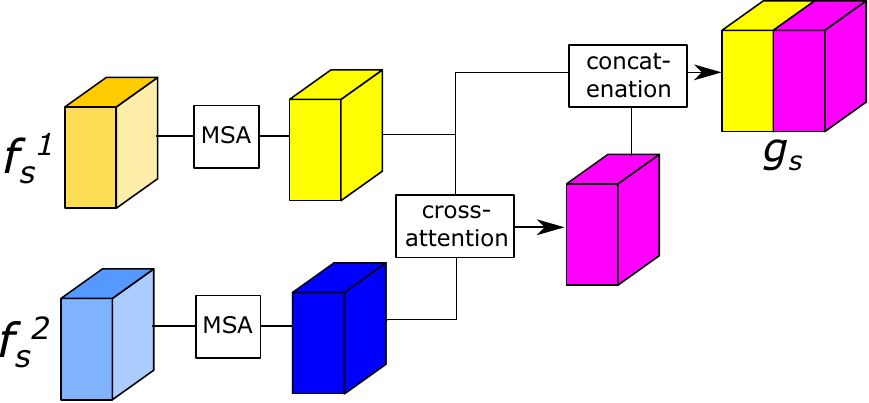}}
    % \hfill
    \hspace{0.5cm}
  \subfloat[Local cross-attention.\label{fig:localCA_vis}]{%
      \includegraphics[width=0.27\linewidth]{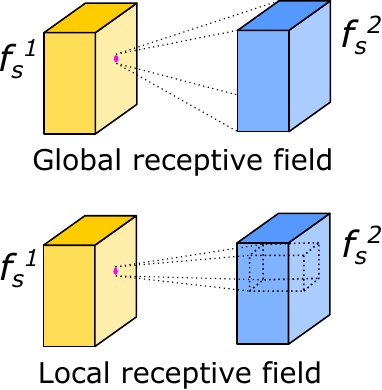}}
  \caption{Two proposed variations to feature registration based on global cross-attention.}
  \label{fig:attention_mechanisms}
\end{figure*}

\subsubsection{Linear multi-headed self-attention:} Firstly, relying on registering each feature vector in $f_s^1$ with (its most similar counterpart in) $f_s^2$ assumes that there is a rich, descriptive and discriminative feature at each spatial location of $f_s^1$. However, assembly images contain a lot of textureless, and therefore featureless, background, which can easily be `matched' by the cross-attention module across images. Thus, if the sample image contains a part that is not in the anchor image, and the spot where this part is missing from the anchor image contains textureless background, it is harder to detect this part as missing from the anchor image. 
Inspired by the work of Sun~\etal on feature matching~\cite{loftr}, we propose to combine GCA with their linear multi-headed self-attention method (GCA+MSA) for enriching features in textureless image areas with global contextual information, demonstrated in~\autoref{fig:MSA+CA}. In theory, this should enable points on textureless regions to be corresponded with the same points across image pairs, where they are described not only by their CNN-extracted local features, but also by the global information from the image. For instance, the global information can contain a point's distance from key assembly object parts. Thus, computing the cross-attention on these globally-enriched feature vectors can enable the model to better identify changes in places where the anchor image contains background pixels.
Instead of full self-attention, we use linear self-attention, since the latter decreases the model's computation complexity at no cost to performance~\cite{loftr}.

\subsubsection{Local cross-attention:}
 Secondly, the cross-attention is global and, although we train with some orientation differences between images, these differences are relatively minor, meaning that the same part in each image pair will have a similar spatial location. Thus, computing the cross-attention across the entire image is redundant and can introduce noise. We address this limitation by introducing a local cross-attention (LCA) network, where each feature vector in~$f_s^1$ is given a restricted receptive field within which to find a matching feature vector in~$f_s^2$. This mechanism is illustrated in \autoref{fig:localCA_vis}. It incorporates an inductive bias derived from the task into the model architecture, since we assume that image pairs contain the assembly object with a similar pose. By searching for changes in a more local way, the model becomes less susceptible to being distracted by features in non-corresponding places. This cannot be guaranteed with the standard GCA mechanism, which is completely invariant to spatial location.

\begin{figure*}[t]
    \centering
    \includegraphics[width=\linewidth]{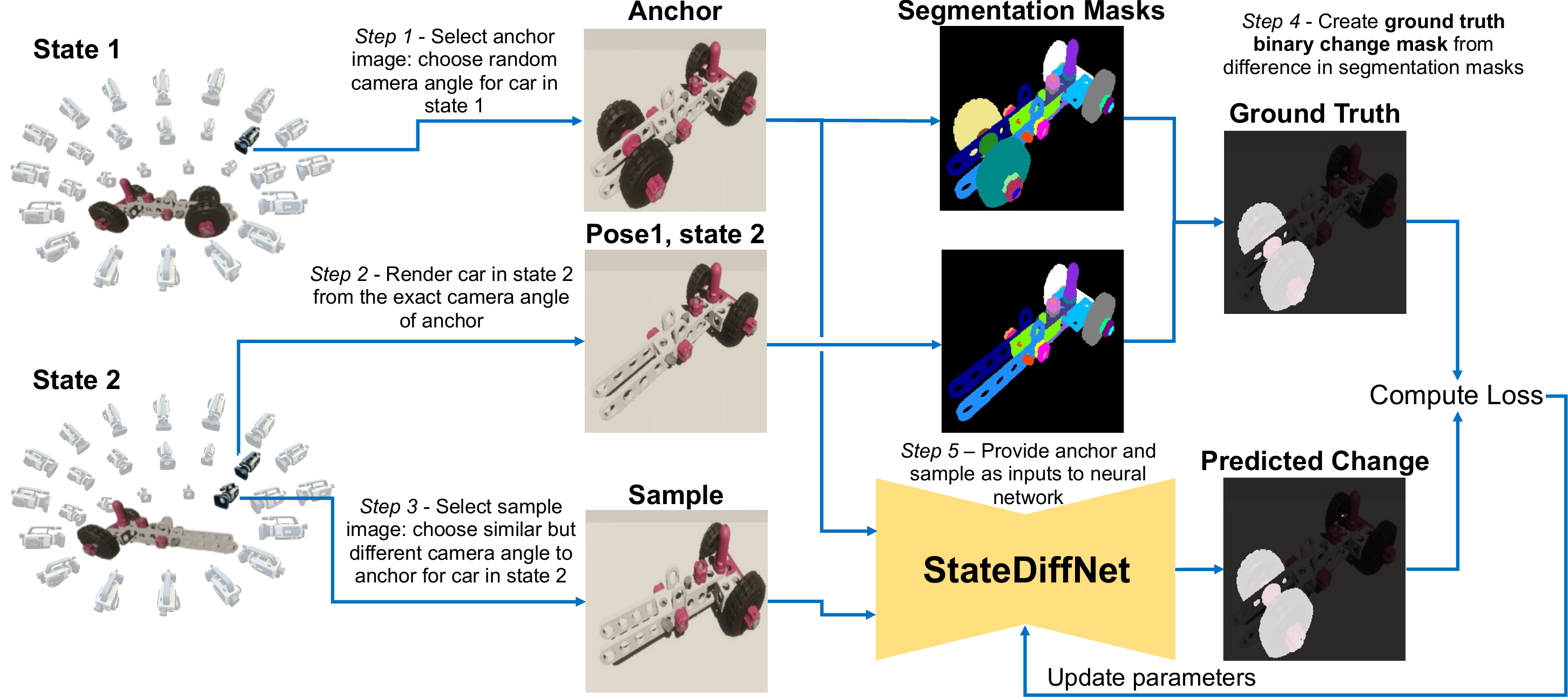}
    \caption{Overview of the training process. The ground-truth binary change mask of an image pair is created by taking the difference between the instance segmentation masks of the anchor and the sample image viewed from the camera angle of the anchor.}
    % \vspace{-0.4cm}
    \label{fig:datagen}
\end{figure*}

\subsection{Data Generation and Image Pair Selection}
\label{sec:data_pipe}
The objective of the proposed system is to capture meaningful change, \eg a missing part, while being resilient to the expected variability between two images. This variability can be caused by, \eg background noise, motion blur and 6D pose differences. Training the model entirely on synthetic data enables full control over both the meaningful change and the expected variability between images. Additionally, it allows a large quantity of training data to be gathered with many randomly selected instances of meaningful change, which we posit to be crucial for a model tasked with finding errors in states that are not in the training data. To this end, we propose an innovative data generation and selection pipeline that can be applied to any assembly object for which a 3D model is available. The process is described in detail in \autoref{fig:datagen}. Briefly, given an assembly object in many different possible states, image pairs are generated by choosing two states with a controlled amount of state difference~(defined by the difference between visible parts), and generating three synthetic images with instance-level segmentation masks: the sample image, the anchor image, and the assembly state of the sample image in the pose of the anchor image. The segmentation masks of the latter two are used to construct a binary change mask, containing all of the meaningful change between both images. We add variance between images by randomizing the background planes and lighting conditions and varying the camera pose. This pipeline provides full control over the object poses, as well as the amount of part-level differences between the anchor and sample image used during training and testing.

\section{Experiments}

This section outlines our experiments. First, we provide the experimental setup and demonstrate that our model performs well at change segmentation on synthetic images with unseen object states, both for object-camera poses seen during training as well as for novel poses. Additionally, a qualitative analysis of the model's performance on real-world test data is provided. To understand the functionality and limitations of the model, an ablation study is presented, where the approach is trained and applied to an even more challenging dataset, wherein the test set contains not just novel part configurations but entirely novel parts.

\subsection{Experimental Setup}

\subsubsection{Datasets:}
In this work, we use the IndustReal~\cite{industreal} dataset. The assembly is performed on a toy car composed of 34 parts that can be assembled following many different orders, generating potentially millions of different states in the proposed synthetic data generation pipeline. The car's pose relative to the camera varies along all six degrees of freedom, pieces of the car to be assembled are scattered in the background, and parts of the car are occluded by the user's hands when being held. Based on the 3D model of the toy car, we generate a training dataset consisting of 10$^6$ images, containing 5,000 different states and 200~camera poses per state, selected to give a top-side view of the car. The data are generated using Unity Perception as a simulator~\cite{unity-perception2022} and the rendered states are created by randomly removing parts without creating disconnected parts. The data generation pipeline is made publicly available.

Two~(synthetic) test sets are used to evaluate StateDiffNet. The first has exactly the same poses as the training data, but contains previously unseen states. The second test set contains the same `unseen' states, but with exclusively novel poses outside the range of the ones in the training set. We qualitatively test the model on real-world images with assembly errors from the IndustReal dataset, including error types that are entirely missing from the training and test sets, namely misplacement or wrong orientation of components. Each real-world error image is paired with a \textit{synthetic} anchor image containing its intended state under a sufficiently similar pose to the real image.

\subsubsection{Implementation details:}
Experiments are conducted using a ResNet-34 encoder~\cite{resnet} backbone with ImageNet-pretrained weights and a two-layer convolutional segmentation head. The cross-attention module is used as skip connection at the spatial resolutions 8$\times$8, 16$\times$16, and 32$\times$32, with channel resolutions~512, 256 and~128, respectively. At each of those resolutions, the GCA+MSA model introduces a linear self-attention layer according to the implementation of LoFTR~\cite{loftr} to both feature maps prior to the cross-attention layer, as shown in \autoref{fig:MSA+CA}. Positional encoding is added to the feature maps prior to the self-attention, which uses 8~heads and a dimension equivalent to each channel resolution. At each spatial resolution, the model trained with LCA use an empirically selected cross-image receptive field of~5$\times$5, 7$\times$7, and~11$\times$11, respectively.

All images are region-of-interest~(ROI) cropped and resized to~256$\times$256 pixels, and color jitter, random rotation, vertical flipping, horizontal flipping and Gaussian blurring are used as image augmentations on the training set. ROI crops are generated by adding a 10\% margin to the bounding-box annotation of the object in each image and randomly translating the object therein. Example training pairs can be viewed in \autoref{sec:synth_training} of the supplementary materials. Each experiment uses a batch size of~64, with 15~warmup epochs and a learning rate of~10$^{-5}$ with cosine learning rate decay to~0 over 400~epochs using the Adam optimizer~\cite{Adam} to minimize the cross-entropy loss. Following the convention for segmentation tasks, we report our results on the Intersection over Union~(IoU) metric~\cite{BIT} over the \textit{change} class. Unless stated otherwise, we set the number of differences in parts to detect during training between~1 and~6 and the threshold on the orientation difference between anchor and sample images to~0.1 norm of quaternion difference (nQD), a suitable metric proposed by Huynh~\cite{huynh-2009}. The reader is kindly referred to \autoref{sec:nQD} of the supplementary materials for further clarification and visualization of the nQD metric.  

\subsection{Error Localization Performance}

\begin{figure}[tb]
\centering
\begin{subfigure}{.5\textwidth}
   \centering
    \includegraphics[width = \linewidth]{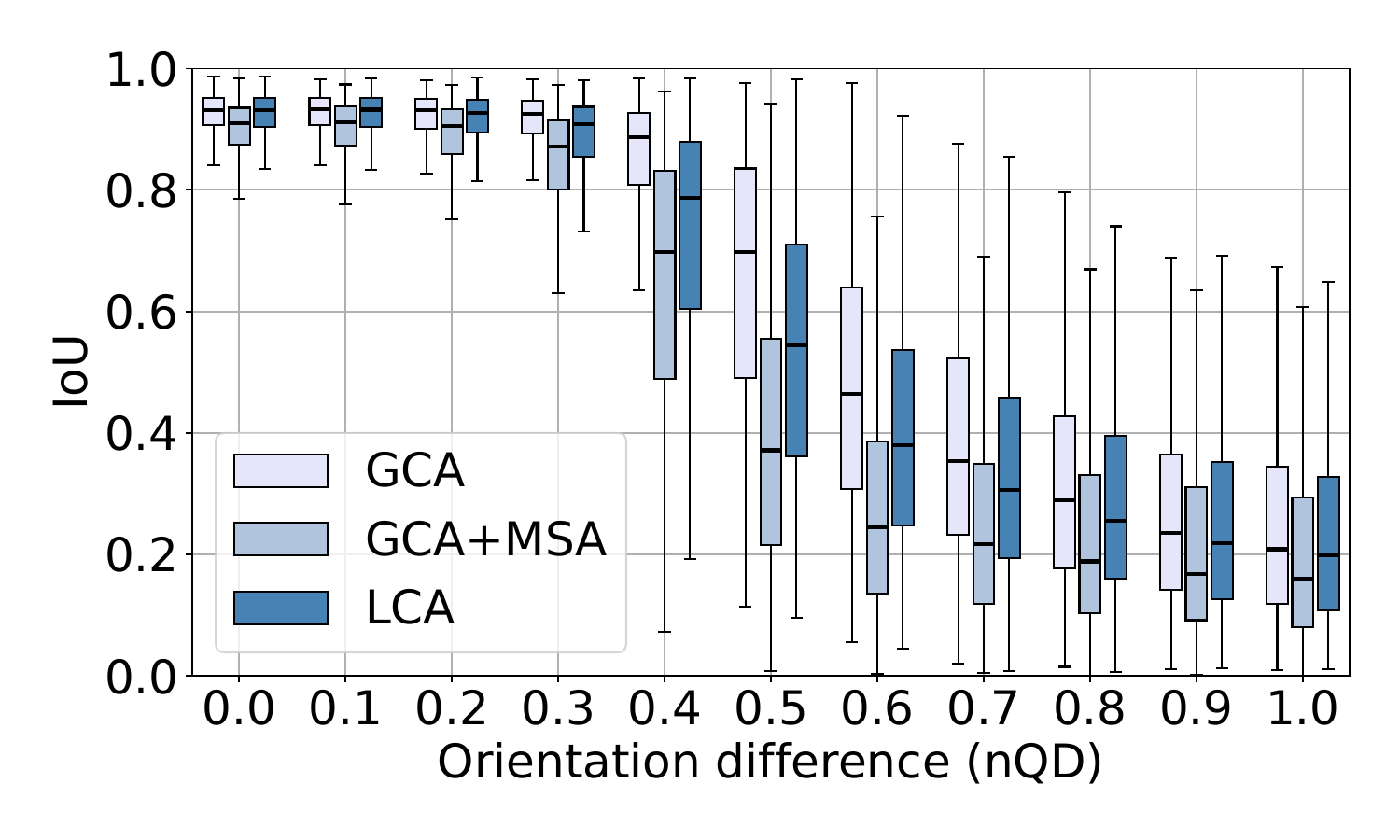}
    \caption{Same poses as training set}
    \label{fig:v2_quant_results}
\end{subfigure}%
\begin{subfigure}{.5\textwidth}
  \centering
    \includegraphics[width = \linewidth]{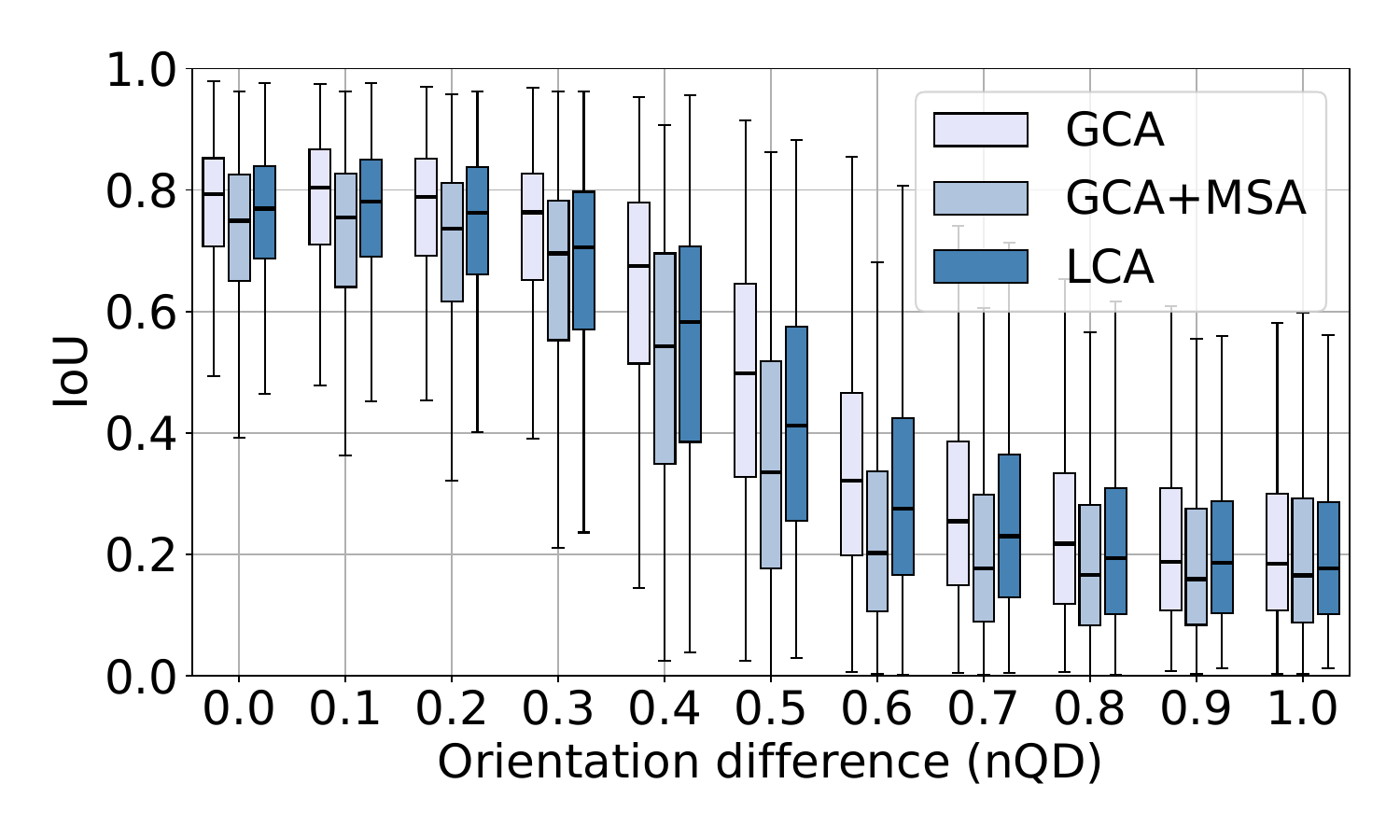}
    \caption{Different poses than training set}
    \label{fig:v2_extra}
\end{subfigure}
\caption{Performance of three models with varying orientation difference. The box plot shows the median and first and third quartiles.}
\label{fig:test}
\end{figure}

\subsubsection{Synthetic test sets:}
The results of all three tested models, each evaluated on varying orientation differences and with 4~to~6~parts as change between the anchor and sample images, are shown in~\autoref{fig:v2_quant_results}. Results for different amounts of part-changes may be found in the supplementary materials,~\autoref{sec:supp_extra_part_diffs}. It can be observed that the model with global cross-attention (GCA) performs best, especially at high orientation differences, and the model with linear MSA performs worst. The local cross-attention performs similarly to the GCA at small orientation differences, but, as expected, degrades for large orientation differences. 

Nevertheless, despite being trained with only relatively small orientation differences (0.1 nQD), most of the models exhibit decent performance on image pairs with considerably more orientation differences, up to 0.3 or 0.4 nQD. When evaluated on poses that are not seen during training, as shown in \autoref{fig:v2_extra}, the overall performance of all models drops by around~0.1 IoU, and the variability in performance between images (whisker size) grows considerably. This indicates that the models are overfitted on the range of poses encountered during training. 
\begin{figure}[tb]
    \centering
    \includegraphics[width=\textwidth]{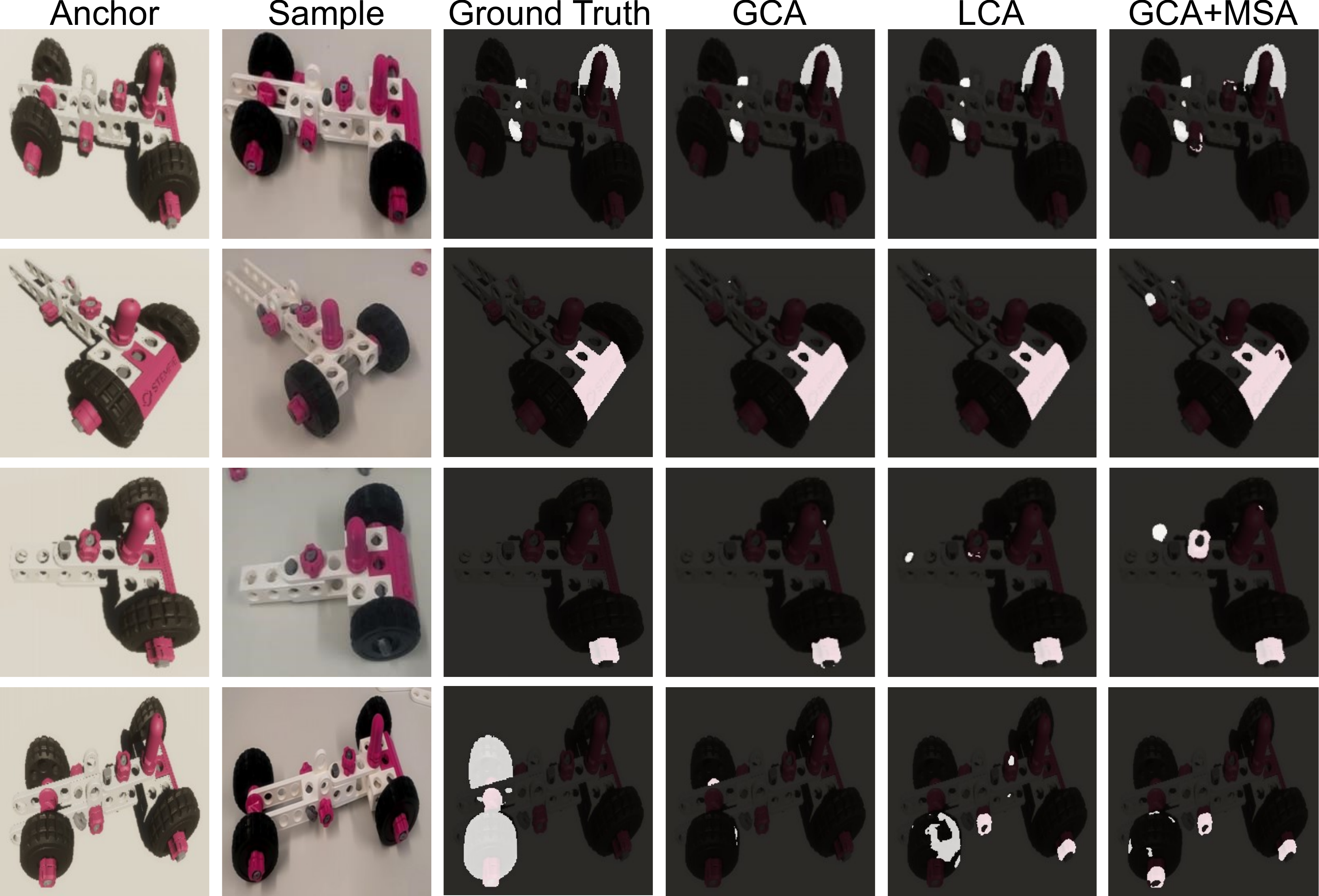}
    \caption{Real-world qualitative results on different assembly errors. The global cross-attention based model performs best, particularly on missing parts, components, and mis-orientations. No model is able to reliably detect placement errors~(last row).
    }
    \label{fig:real_world_results}
\end{figure}
\subsubsection{Real-world test set:}
The qualitative results of all three models on real-world images, without using any explicit domain adaptation techniques, are shown in \autoref{fig:real_world_results}. All models exhibit decent performance at detecting the assembly state errors, despite the obvious differences in visual appearance between real and synthetic images. The GCA model has the most accurate predictions, particularly because it scores less false positives compared to the other attention mechanisms. The qualitative analysis suggests that our approach to assembly error localization is able to handle missing~(Rows~1-2) and mis-oriented parts~(Row~3), but fails at localizing placement errors~(Row~4), where correct components are assembled in the wrong place. Furthermore, the models successfully disregard scattered background objects that do not constitute meaningful change (Rows~2-4).
\subsection{Ablation Study}
In addition to demonstrating that StateDiffNet is capable of generalizing to new configurations, poses, and real-world data, we perform an ablation study. To this end, we investigate (1)~whether the proposed system can be used out-of-the-box in scenarios where a part of the assembly is upgraded, \ie how it performs on parts that were never seen as change during training, (2)~the impact of orientation differences between the anchor and sample in training, and (3)~the difference between a part being added versus removed between the anchor and sample. This ablation aims to investigate the GCA performance on a more complex task and to understand the limitations of the cross-attention mechanism. 

For the new objective of detecting change in parts that are entirely unseen and/or never changed during training, a new training set is created, consisting of~1,000 different states of the car with the same~200 camera poses per state as used previously. The dataset is designed such that one part, specifically the front bracket, is always present, while two parts~(the pulley and fourth wheel) are never present. Therefore, no change segmentation is ever obtained on those three parts during training. The same synthetic test set, which contains no limitations on the parts that are removed, is used as for prior experiments.

\subsubsection{Perfect image alignment versus slight orientation differences:} 
\begin{figure}[tb]
    \centering
    \includegraphics[width=0.99\linewidth]{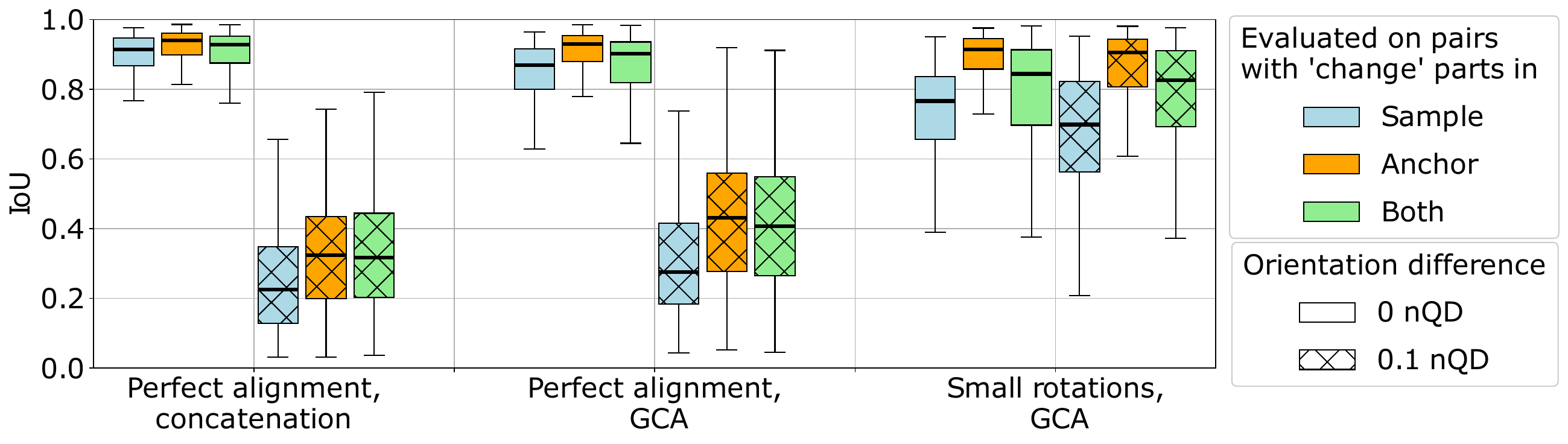}
    \caption{Performance on perfectly aligned and differently oriented images for three different models. The first two are trained on aligned image pairs, one with simple feature concatenation, the other using cross-attention. The third is trained on slight orientation differences. The colors represent the origin of the changes to be segmented, \ie parts that are only present in the sample~(blue), anchor~(orange) or both~(green) images, and the hatches indicate the maximum nQD encountered during testing.}
    \label{fig:missing_vs_present}
    % \vspace{-0.6cm}
\end{figure}

To validate the fundamental feasibility of a model that generalizes to detecting changes on unseen parts, we first test whether this is possible for two images that are perfectly registered. This is tested by training one StateDiffNet without any attention mechanism, by simply concatenating the two feature maps instead of using cross-attention, and another StateDiffNet with global cross-attention, on perfectly aligned images. The results are presented in \autoref{fig:missing_vs_present}, and demonstrate that the best performance is achieved when cross-attention is not used. This is intuitive, since the images are already perfectly aligned, rendering the use of cross-attention to register features across images obsolete and more sensitive to noise. Since features only need to be compared locally, a purely convolutional network performs best at detecting change on registered images. The network can even detect changes never encountered during training, as demonstrated in the supplementary materials, \autoref{sec:supp_synth_qualitative}, by simply determining whether identical features are present in both images at each spatial location. We posit that this allows change to be detected at the `feature' level, instead of requiring a semantic understanding of the assembly object to reliably detect change, thereby greatly simplifying the problem and enabling better generalization to unseen parts.

However, as soon as slight pose differences of the assembly object are introduced, the aforementioned models fail to meaningfully detect change, as can be seen in \autoref{fig:missing_vs_present}. Only when the model with GCA is trained on image pairs with orientation differences is it able to segment change on non-aligned images. However, we observe that this model performs worse than both the previous models on perfectly aligned images. This performance drop is caused by the model failing to generalize to new changes when input images are not perfectly aligned. This failure to generalize is clearly visible in the qualitative results, for which we refer the reader to \autoref{sec:supp_synth_qualitative} of the supplementary materials. 

In conclusion, change detection without attention generalizes well to change in unseen parts when images are perfectly aligned. However, when the images contain some orientation difference, cross-attention is required for feature registration, removing the model's ability to generalize to this type of novel change.%
\subsubsection{Parts missing in anchor versus parts missing in sample:} 
StateDiffNet is tasked with segmenting all the changes identified between image pairs on the \textit{anchor} image. Thus, for changes of parts that are present in the anchor and missing from the sample image, the model needs to predict a segmentation mask over those parts. However, when the change is caused by a part missing from the anchor image, the model needs to provide a segmentation mask corresponding to where that part \textit{would be in the anchor image}, which is a considerably more difficult task. The increased difficulty is highlighted in \autoref{fig:missing_vs_present}. Regardless of whether the model is trained on perfectly aligned images, it performs best for parts present in the anchor image. This performance difference is amplified when the model is trained on non-aligned input images: the model's segmentation performance of parts present in the anchor image drops by~2\% IoU when introducing orientation change, whereas it drops by~10\% when the parts are only present in the sample image. This highlights a fundamental flaw of the cross-attention mechanism: features from the sample image are only transferred to the decoder if similar features are present in the anchor. Therefore, features from the sample image that do not match any features from the anchor image are lost.

\section{Discussion}
\subsubsection{Global cross-attention mechanism:}
\begin{figure*}[tb]
  \centering
  \subfloat[Region unchanged between images.\label{fig:att_both}]{%
       \includegraphics[width=0.24\linewidth]{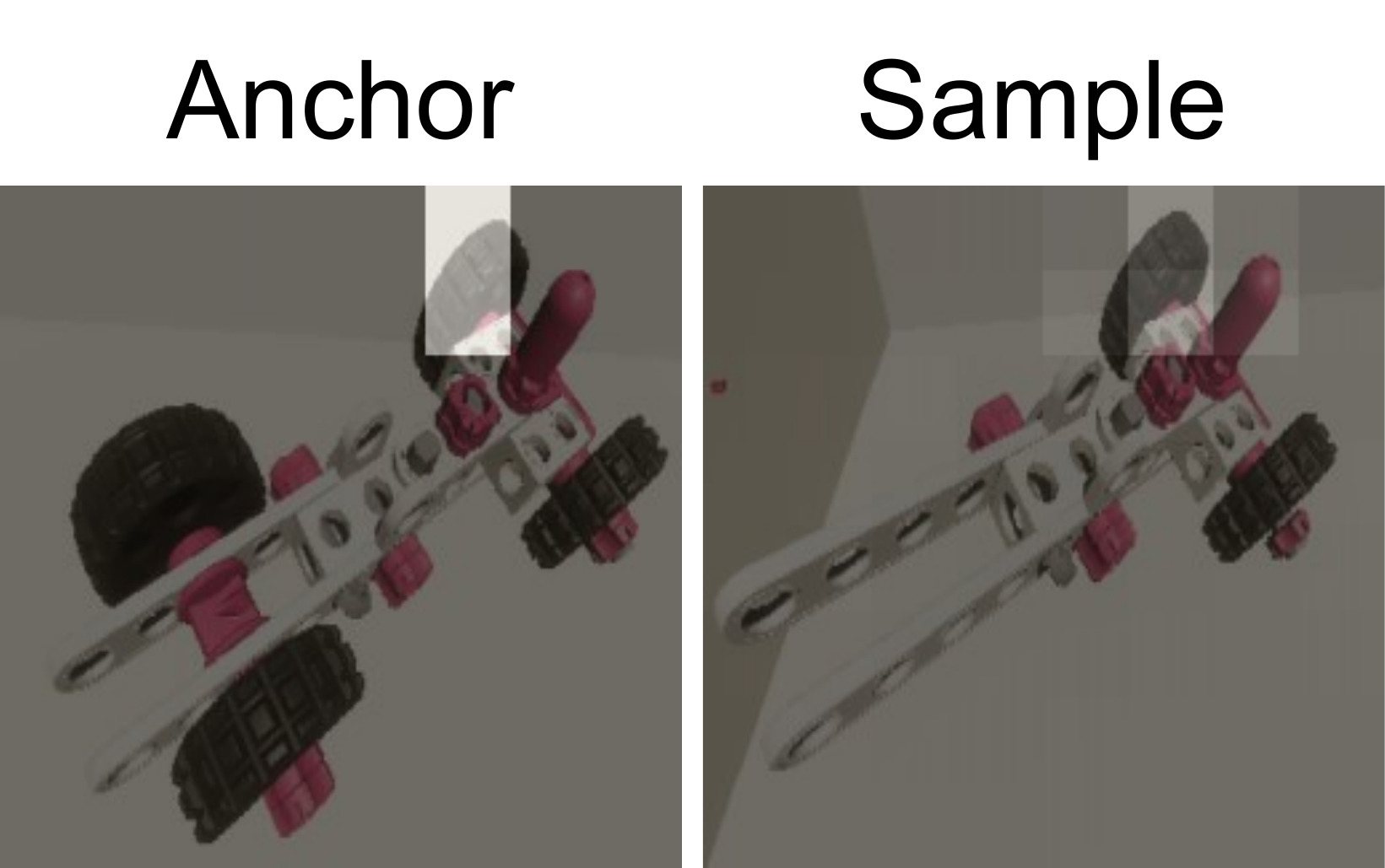}}
    \hfill
  \subfloat[Background region of image queried.\label{fig:att_background}]{%
      \includegraphics[width=0.24\linewidth]{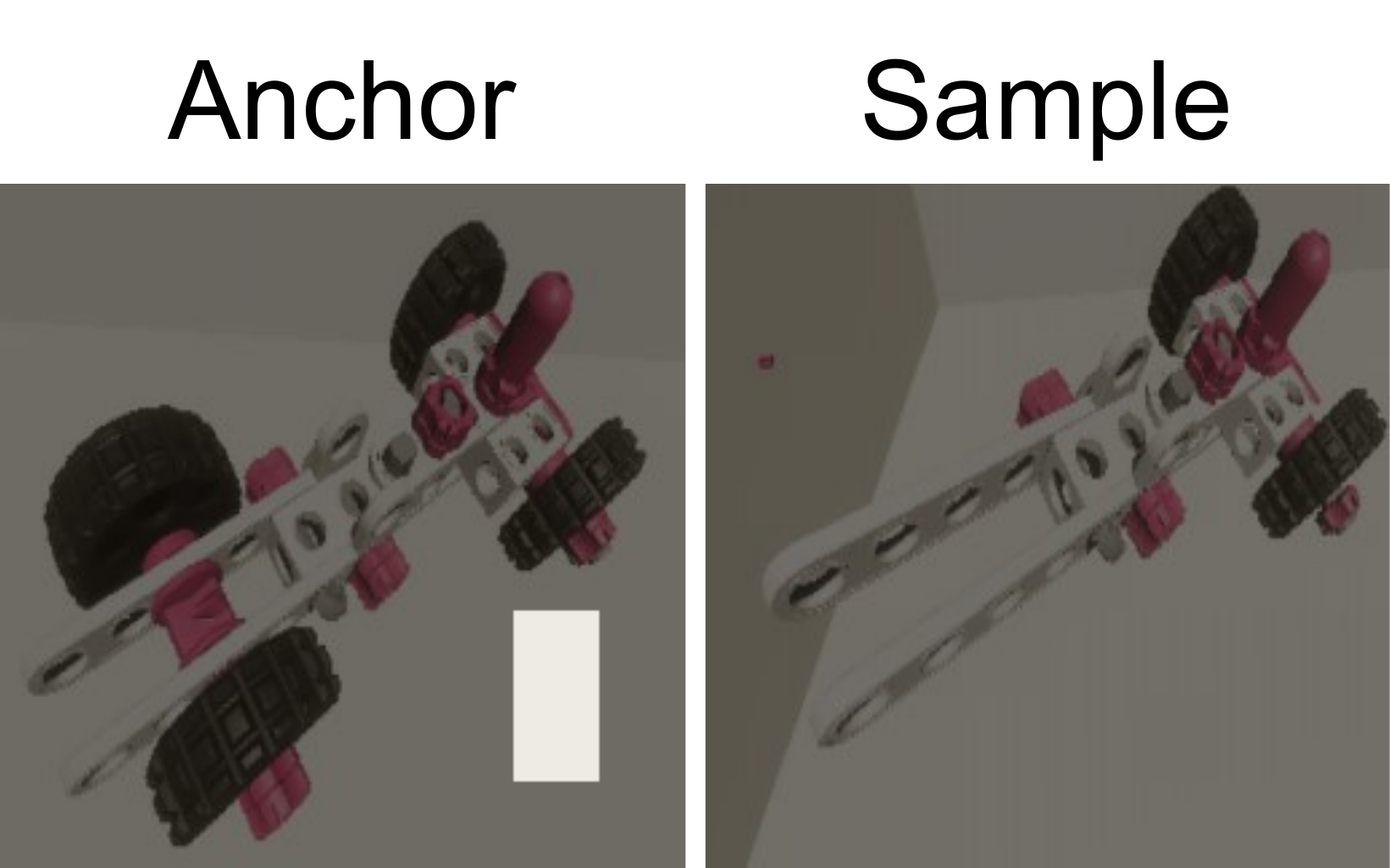}}
  \hfill
  \subfloat[Region removed in sample image.\label{fig:att_removed}]{%
        \includegraphics[width=0.24\linewidth]{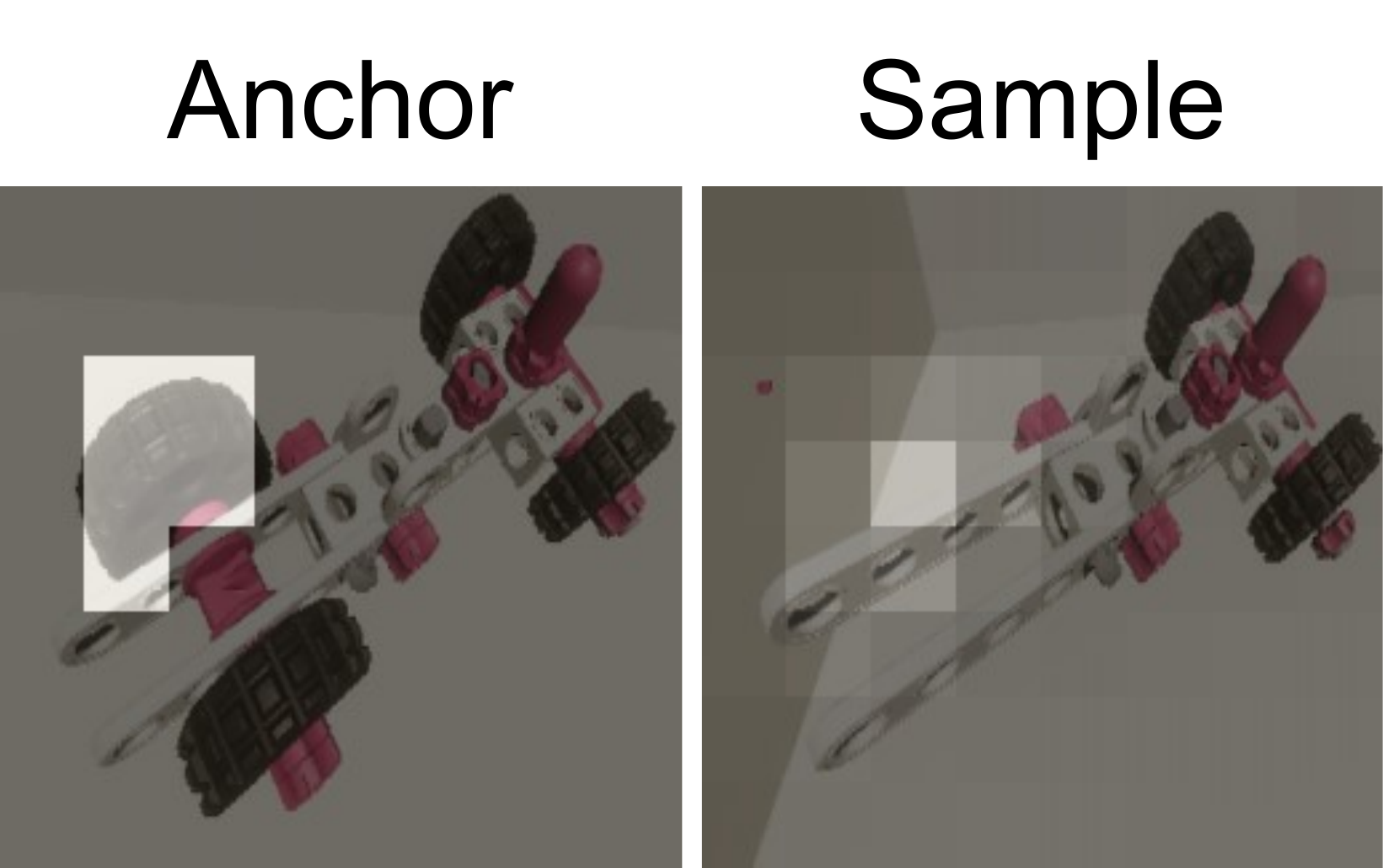}}
    \hfill
  \subfloat[Region added in sample image.\label{fig:att_added}]{%
        \includegraphics[width=0.24\linewidth]{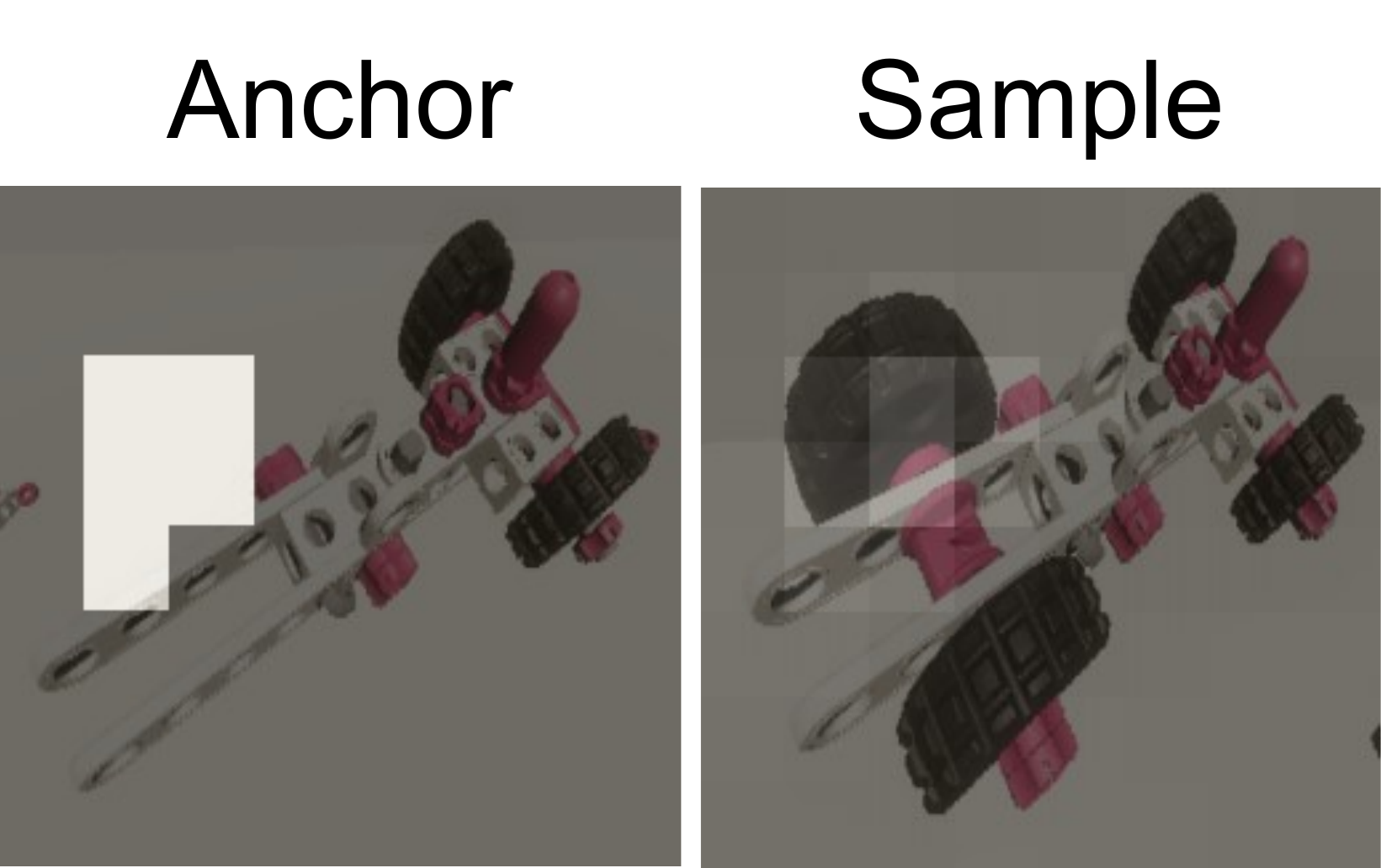}}
    \hfill
  \caption{Visualization of GCA weights on a sample image for queried areas on anchor image. The GCA behaves as expected by attending to the most similar region in the sample (a). Similarly, GCA behaves as expected by not attending to any specific region when the queried background region matches many regions in the sample image (b). However, for parts present in the anchor but not in the sample~(c), the GCA attends to the region where the part is expected to be located in the sample. Surprisingly, when selecting a background region in the anchor image where a part can potentially be placed~(d), the GCA attends to the region in the sample image where this part is actually placed. Best viewed on a computer monitor.}
  \label{fig:attention_weights}
\end{figure*}

This work has compared three attention mechanisms for feature registration, namely global cross-attention (GCA), local cross-attention~(LCA), and multi-headed self-attention~(GCA+MSA). These three mechanisms are compared, because it is not intuitive that GCA is suitable for error localization: it lacks positional encoding and intuitively \textit{should} therefore not be able to register features from feature-less regions. However, the results surprisingly demonstrate that the GCA mechanism performs best. To understand this, we visualize the cross-attention weights of the GCA in \autoref{fig:attention_weights}. 

The example in \autoref{fig:attention_weights} demonstrates that, even if the queried area consists of mostly background, the GCA attends to the wheel in the sample image that is missing from the queried area in the anchor~(\autoref{fig:att_added}). Therefore, we conclude that the cross-attention does not match visually similar features as hypothesized and shown by~\cite{cyws} on their change detection dataset. Rather surprisingly, despite having no spatial encoding of the tokens in the cross-attention, it is apparently capable of leveraging contextual information~(\eg the parts surrounding the wheel) to find the matching wheel in the sample image. This is further evidence that, at least in our experimental setup, the cross-attention has semantic understanding of the parts and their 3D configuration, and does not merely operate at a feature-matching and registration level.

\subsubsection{Memorization versus generalization:}
The tested models all suffer a performance drop for poses that were never encountered during training, compared to poses that lie within the training set~(\autoref{fig:v2_extra}). This result suggests that the models may determine whether all parts in one image are present in the other, learn missing part locations for each pose by heart, and produce the memorized segmentation mask for any parts that are changed. This is further evidenced by the worse generalization to entirely new parts when that part is missing in the anchor, compared to when that part is missing from the sample image~(\autoref{fig:missing_vs_present}). Therefore, the proposed algorithm must be retrained when a part is modified or added to the assembly. The model only depends on synthetic data and, therefore, retraining the model only incurs computational cost~(no manual annotation efforts are required). Nonetheless, these insights highlight a valuable direction for future research into the use of cross-attention for change detection.

\subsubsection{Change segmentation versus change detection:}
In this work, error localization is addressed as a segmentation rather than bounding-box detection problem because segmentation provides a more fine-grained prediction. Therefore, segmentation can provide more feedback to operators, creating a better user interface, \eg for an AR system. Nonetheless, training for segmentation rather than bounding-box prediction has the inherent flaw that the model optimizes towards precise segmentation of fine edges, forcing the model to learn the projection of missing parts by heart at the cost of generalization. To improve generalization to new parts and poses while maintaining fine-grained predictions, it can be beneficial to optimize a cost function that less focuses on edges.

\section{Conclusion}
This work discusses error localization for assembly procedures. Anomaly detection approaches are not directly applicable, since the correctness of an object state depends on the intended state in the procedure. Assembly state detection approaches focus on assembly configurations with limited complexity and do not investigate error localization. Therefore, we propose an approach called StateDiffNet, tasked with change segmentation between an image containing the intended assembly state and a sample image. Since the intended states are known to be error-free, any change with respect to the sample image can be considered an error. The algorithm is trained entirely on synthetic data, enabled through a novel data generation pipeline that we make publicly available. This pipeline allows for the explicit modeling of the expected variability between image pairs, such as camera pose, photometry, and image distortions. StateDiffNet demonstrates strong performance for error segmentation on a complex assembly configuration, as shown quantitatively on synthetic data and qualitatively on real-world data. The approach is able to localize errors in unseen assembly configurations and even for entirely unseen error types.

The conducted experiments have surprisingly demonstrated that the cross-attention feature-registration mechanism is capable of leveraging contextual information to attend to feature-less regions. This is unexpected since the mechanism does not have any spatial encoding. Furthermore, we demonstrate that feature registration is the limiting factor in training a model to localize errors on entirely unseen parts. Therefore, we argue that superior feature-registration methods are required, and it is demonstrated that two ideas towards such a method do not result in improved performance. This highlights the clear challenges of employing existing algorithms to the industrial domain and provides various interesting directions for future research.

\section*{Acknowledgment}
The authors express their gratitude to Hans Onvlee for his valuable insights. This work is partially executed at ASML Research, with funding from ASML and TKI grant number TKI2112P07.

\bibliographystyle{splncs04}
\bibliography{main}

\begin{thebibliography}{10}
\providecommand{\url}[1]{\texttt{#1}}
\providecommand{\urlprefix}{URL }
\providecommand{\doi}[1]{https://doi.org/#1}

\bibitem{TinyCD}
A.~Codegoni, G.~Lombardi, A.F.: Tinycd: a (not so) deep learning model for change detection. Neural Computing and Applications  \textbf{35},  8471--8486 (2023). \doi{https://doi.org/10.1007/s00521-022-08122-3}

\bibitem{albers2023augmented}
Albers, A., Bohn{\'e}, T., Tadeja, S.K.: Augmented reality for quality inspection: A user-centred systematic review of use cases, trends and technology. In: 2023 IEEE International Symposium on Mixed and Augmented Reality Adjunct (ISMAR-Adjunct). pp. 146--153. IEEE (2023)

\bibitem{vision-datasets}
Bai, H., Mou, S., Likhomanenko, T., Cinbis, R.G., Tuzel, O., Huang, P., Shan, J., Shi, J., Cao, M.: Vision datasets: A benchmark for vision-based industrial inspection. arXiv preprint arXiv:2306.07890  (2023)

\bibitem{ChangeFormer}
Bandara, W.G.C., Patel, V.M.: A transformer-based siamese network for change detection. In: IGARSS 2022 - 2022 IEEE International Geoscience and Remote Sensing Symposium. pp. 207--210 (2022). \doi{10.1109/IGARSS46834.2022.9883686}

\bibitem{DINO}
Caron, M., Touvron, H., Misra, I., Jégou, H., Mairal, J., Bojanowski, P., Joulin, A.: Emerging properties in self-supervised vision transformers. 2021 IEEE/CVF International Conference on Computer Vision (ICCV) pp. 9630--9640 (2021), \url{https://api.semanticscholar.org/CorpusID:233444273}

\bibitem{AssCD_real}
Chen, C., Yue, Y., Wang, J.: Multi-view change detection method for mechanical assembly images based on feature fusion and feature refinement with depthwise separable convolution. Multimedia Tools and Applications  (07 2023). \doi{10.1007/s11042-023-16165-4}

\bibitem{prego}
Flaborea, A., di~Melendugno, G.M.D., Plini, L., Scofano, L., Matteis, E.D., Furnari, A., Farinella, G.M., Galasso, F.: Prego: online mistake detection in procedural egocentric videos (2024)

\bibitem{BIT}
Hao~Chen, Z.Q., Shi, Z.: Remote sensing image change detection with transformers. IEEE Transactions on Geoscience and Remote Sensing pp. 1--14 (2021). \doi{10.1109/TGRS.2021.3095166}

\bibitem{resnet}
He, K., Zhang, X., Ren, S., Sun, J.: Deep residual learning for image recognition. In: 2016 IEEE Conference on Computer Vision and Pattern Recognition (CVPR). pp. 770--778 (2016). \doi{10.1109/CVPR.2016.90}

\bibitem{huynh-2009}
Huynh, D.Q.: {Metrics for 3D Rotations: Comparison and analysis}. Journal of mathematical imaging and vision  \textbf{35}(2),  155--164 (6 2009). \doi{10.1007/s10851-009-0161-2}, \url{https://doi.org/10.1007/s10851-009-0161-2}

\bibitem{reconpatch}
Hyun, J., Kim, S., Jeon, G., Kim, S.H., Bae, K., Kang, B.J.: Reconpatch: Contrastive patch representation learning for industrial anomaly detection. In: Proceedings of the IEEE/CVF Winter Conference on Applications of Computer Vision. pp. 2052--2061 (2024)

\bibitem{spot-the-diff_dataset}
Jhamtani, H., Berg-Kirkpatrick, T.: Learning to describe differences between pairs of similar images. In: Riloff, E., Chiang, D., Hockenmaier, J., Tsujii, J. (eds.) Proceedings of the 2018 Conference on Empirical Methods in Natural Language Processing. pp. 4024--4034. Association for Computational Linguistics, Brussels, Belgium (Oct-Nov 2018). \doi{10.18653/v1/D18-1436}, \url{https://aclanthology.org/D18-1436}

\bibitem{cd_survey}
Jiang, H., Peng, M., Zhong, Y., Xie, H., Hao, Z., Lin, J., Ma, X., Hu, X.: A survey on deep learning-based change detection from high-resolution remote sensing images. Remote Sensing  \textbf{14}(7) (2022). \doi{10.3390/rs14071552}, \url{https://www.mdpi.com/2072-4292/14/7/1552}

\bibitem{Adam}
Kingma, D., Ba, J.: Adam: A method for stochastic optimization. In: International Conference on Learning Representations (ICLR). San Diega, CA, USA (2015)

\bibitem{IED}
Klomp, S., van~de Wouw, D., With, P.: Real-time small-object change detection from ground vehicles using a siamese convolutional neural network. Journal of Imaging Science and Technology  \textbf{63} (11 2019). \doi{10.2352/J.ImagingSci.Technol.2019.63.6.060402}

\bibitem{li-2024}
Li, W., Aibo, X., Wei, M., Zuo, W., Li, R.: {Deep learning-based augmented reality work instruction assistance system for complex manual assembly}. Journal of manufacturing systems  \textbf{73},  307--319 (4 2024). \doi{10.1016/j.jmsy.2024.02.009}, \url{https://www.sciencedirect.com/science/article/pii/S0278612524000323?casa_token=PPUvSttkHqQAAAAA:AMMlr6mQNofosIi_713RYVDhGPMyBJztpsF2CCqFxzLwLtBQ7voV92WpTSCcx2tu-kSwjgzWXw}

\bibitem{unet}
Ronneberger, O., Fischer, P., Brox, T.: U-net: Convolutional networks for biomedical image segmentation. In: Navab, N., Hornegger, J., Wells, W.M., Frangi, A.F. (eds.) Medical Image Computing and Computer-Assisted Intervention -- MICCAI 2015. pp. 234--241. Springer International Publishing, Cham (2015)

\bibitem{cyws}
Sachdeva, R., Zisserman, A.: The change you want to see. In: Proceedings of the IEEE/CVF Winter Conference on Applications of Computer Vision (WACV) (2023)

\bibitem{cyws3d}
Sachdeva, R., Zisserman, A.: The change you want to see (now in 3d). In: Proceedings of the IEEE/CVF International Conference on Computer Vision (ICCV) (2023)

\bibitem{jst2015change}
Sakurada, K., Okatani, T.: Change detection from a street image pair using cnn features and superpixel segmentation  (2015)

\bibitem{asdf}
Schieber, H., Li, S., Corell, N., Beckerle, P., Kreimeier, J., Roth, D.: Asdf: Assembly state detection utilizing late fusion by integrating 6d pose estimation (2024)

\bibitem{simstate}
Schoonbeek, T.J., Balachandran, G., Onvlee, H., Houben, T., et~al.: Supervised representation learning towards generalizable assembly state recognition. In: arXiv preprint arXiv:2408.11700 (2024). \doi{10.48550/arXiv.2408.11700}

\bibitem{industreal}
Schoonbeek, T.J., Houben, T., Onvlee, H., van~der Sommen, F., et~al.: Industreal: A dataset for procedure step recognition handling execution errors in egocentric videos in an industrial-like setting. In: Proceedings of the IEEE/CVF Winter Conference on Applications of Computer Vision. pp. 4365--4374 (2024)

\bibitem{state_aware_detection}
Stanescu, A., Mohr, P., Kozinski, M., Mori, S., Schmalstieg, D., Kalkofen, D.: State-aware configuration detection for augmented reality step-by-step tutorials. In: 2023 IEEE International Symposium on Mixed and Augmented Reality (ISMAR). pp. 157--166 (2023). \doi{10.1109/ISMAR59233.2023.00030}

\bibitem{coffeeMachine}
Su, Y., Rambach, J., Minaskan, N., Lesur, P., Pagani, A., Stricker, D.: Deep multi-state object pose estimation for augmented reality assembly (08 2019). \doi{10.1109/ISMAR-Adjunct.2019.00-42}

\bibitem{loftr}
Sun, J., Shen, Z., Wang, Y., Bao, H., Zhou, X.: {LoFTR}: Detector-free local feature matching with transformers. CVPR  (2021)

\bibitem{unity-perception2022}
{Unity Technologies}: Unity {P}erception package. \url{https://github.com/Unity-Technologies/com.unity.perception} (2020)

\bibitem{AssCD_synth}
Wu, S., Chen, C., Wang, J.: Mechanical assembly monitoring method based on semi-supervised semantic segmentation. Applied Sciences  \textbf{13}(2) (2023). \doi{10.3390/app13021182}, \url{https://www.mdpi.com/2076-3417/13/2/1182}

\end{thebibliography}

\newpage
\section*{Supplementary materials}

\appendix
This appendix consists of supplementary materials that show additional results and provide details that were omitted from the main text for conciseness.

\FloatBarrier 

\section{Quantification of orientation difference} \label{sec:nQD}

\subsubsection{Definition:} The norm of the difference of quaternions (nQD) is defined as 
\begin{equation}
    \Phi(\boldsymbol{q_1},\boldsymbol{q_2}) = \text{min}\{||\boldsymbol{q_1} - \boldsymbol{q_2}||, ||\boldsymbol{q_1} + \boldsymbol{q_2} ||\},
\end{equation}
where $||\cdot||$ denotes the Euclidean norm and $\boldsymbol{q_1}$ and $\boldsymbol{q_2}$ are unit quaternions. The minimum operation is required because unit quaternions $q$ and $-q$ denote the same rotation.

\subsection{Motivation:} The quantification of orientation difference is used in this work to systematically have control over the amount of viewpoint variance between image pairs presented to the model during training and testing. Huynh \cite{huynh-2009} compares multiple potential metrics to quantify rotation difference. The norm of quaternion difference is found to be a suitable metric as small values correspond to nearby rotations and large values correspond to distant rotations (unlike eg. the euclidean distance between Euler angles). Moreover, we choose the nQD metric over some of the other viable options presented in \cite{huynh-2009} as it is more intuitive and simpler to implement. 

\subsection{Illustration}
\begin{figure}
    \centering
    \includegraphics[width=\linewidth]{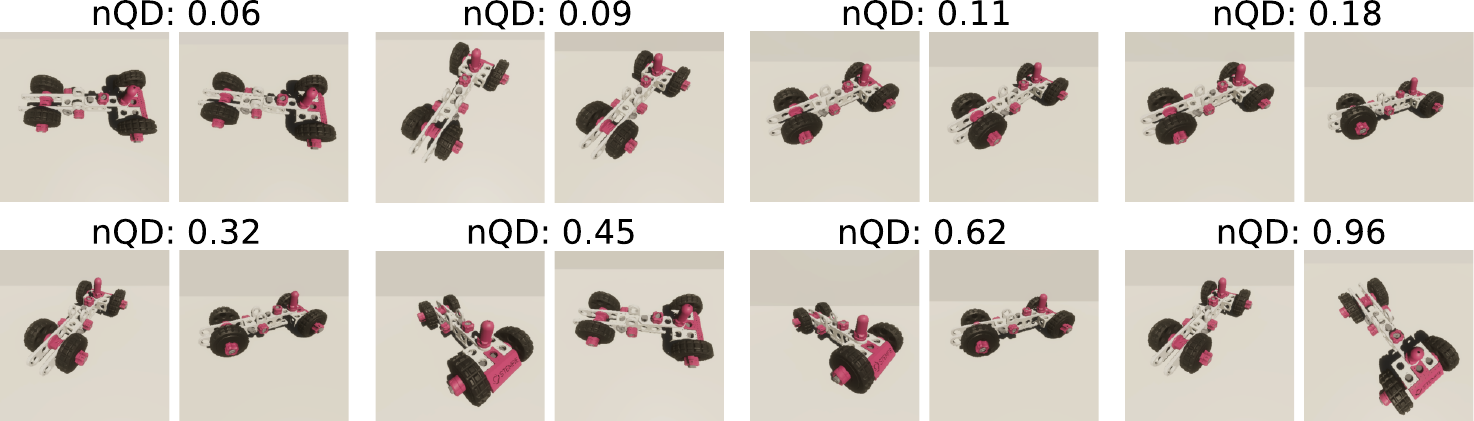}
    \caption{Visualization of the norm of quaternion difference (nQD) measure for orientation difference between various image pairs. }
    \label{fig:nQD_vis}
\end{figure}

To provide a feel for the unit of nQD, we show some example of image pairs and their norm of quaternion difference in \autoref{fig:nQD_vis}. It can clearly be seen that an increase in nQD indeed corresponds to a greater observable rotation difference between two images. Note that the range of possible nQD values lies between 0 and $\sqrt{2} \approx 1.4$.

\section{Training examples}
\label{sec:synth_training}

\begin{figure}
    \centering
    \includegraphics[width=\textwidth]{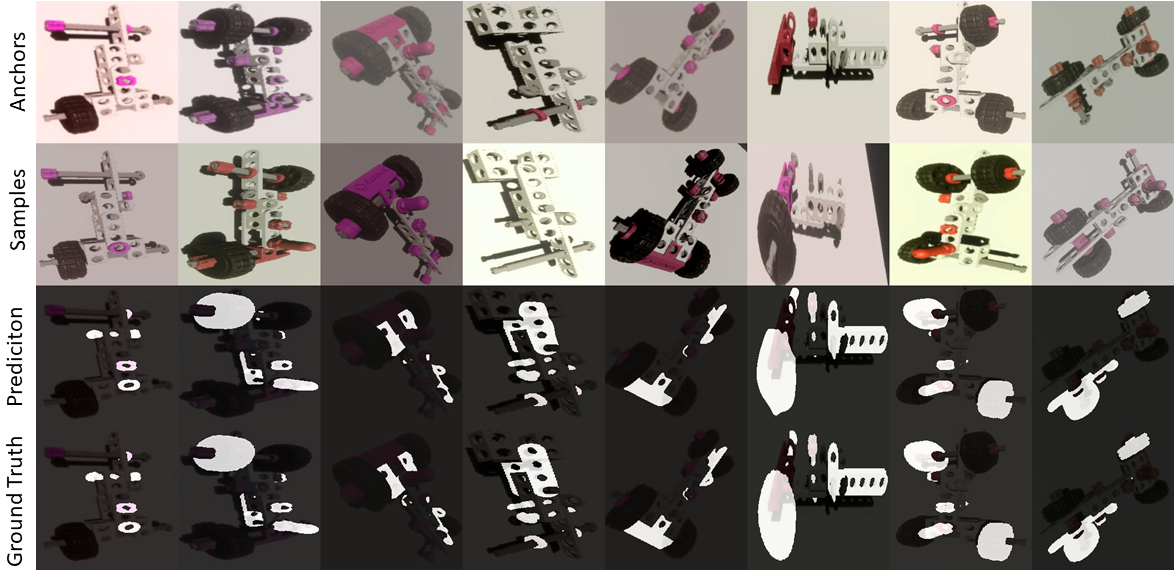}
    \caption{Visualization of some training pairs with the used data augmentations, and the global cross-attention model's predictions}
    \label{fig:training_examples}
\end{figure}

\autoref{fig:training_examples} contains random samples from a batch of training image pairs, along with their ground truth change masks and the predicted change masks generated by the GCA model during training. The image is intended to provide a feel for the kind and amount of image augmentation used on training pairs. It can be seen that there is significant hue, brightness and contrast jitter, used as a `domain generalization' technique. These image augmentations enable the model to perform decently on real-world images without using any domain adaptation techniques.

\FloatBarrier 

\section{Quantitative and qualitative results on synthetic anchor-sample pairs}
\label{sec:supp_extra_part_diffs}

\subsubsection{Quantitative:} 
\begin{figure}[]
\centering
\begin{subfigure}{.5\textwidth}
   \centering
    \includegraphics[width = \linewidth]{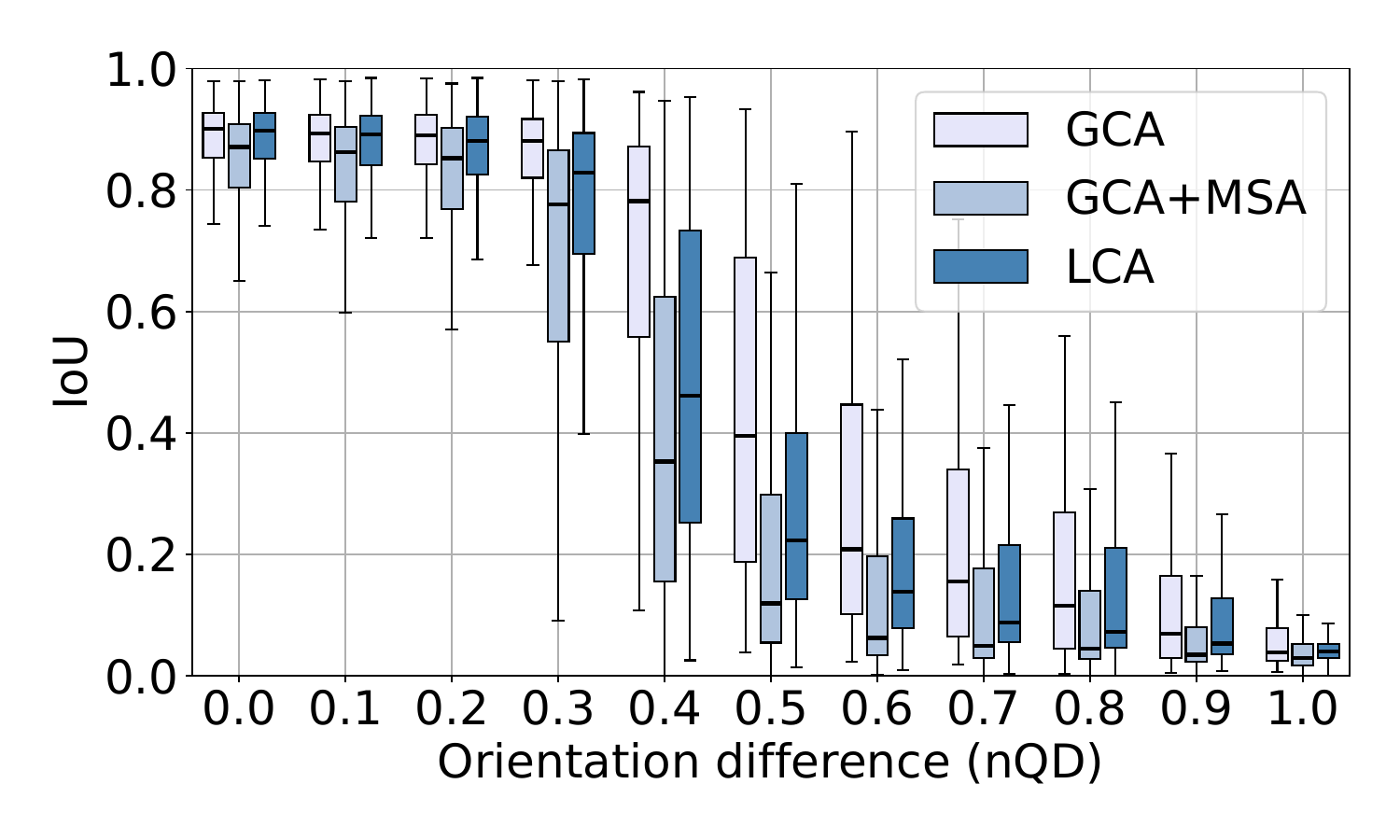}
    \caption{Same poses as training set, 1-3 parts difference between anchor and sample.}
    \label{fig:v2_quant_results_3part}
\end{subfigure}%
\begin{subfigure}{.5\textwidth}
  \centering
    \includegraphics[width = \linewidth]{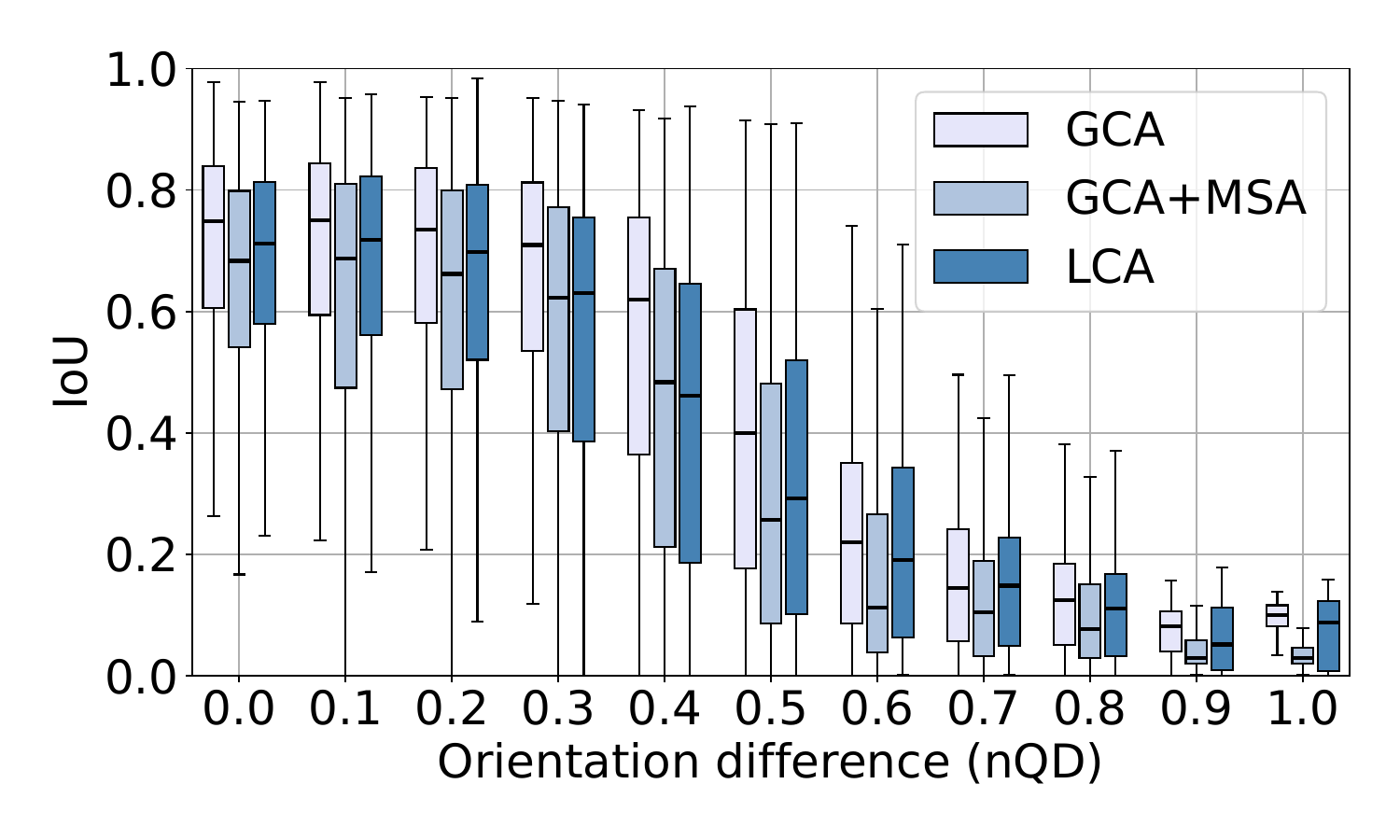}
    \caption{Different poses than training set, 1-3 parts difference between anchor and sample.}
    \label{fig:v2_extra_3part}
\end{subfigure}
\begin{subfigure}{.5\textwidth}
   \centering
    \includegraphics[width = \linewidth]{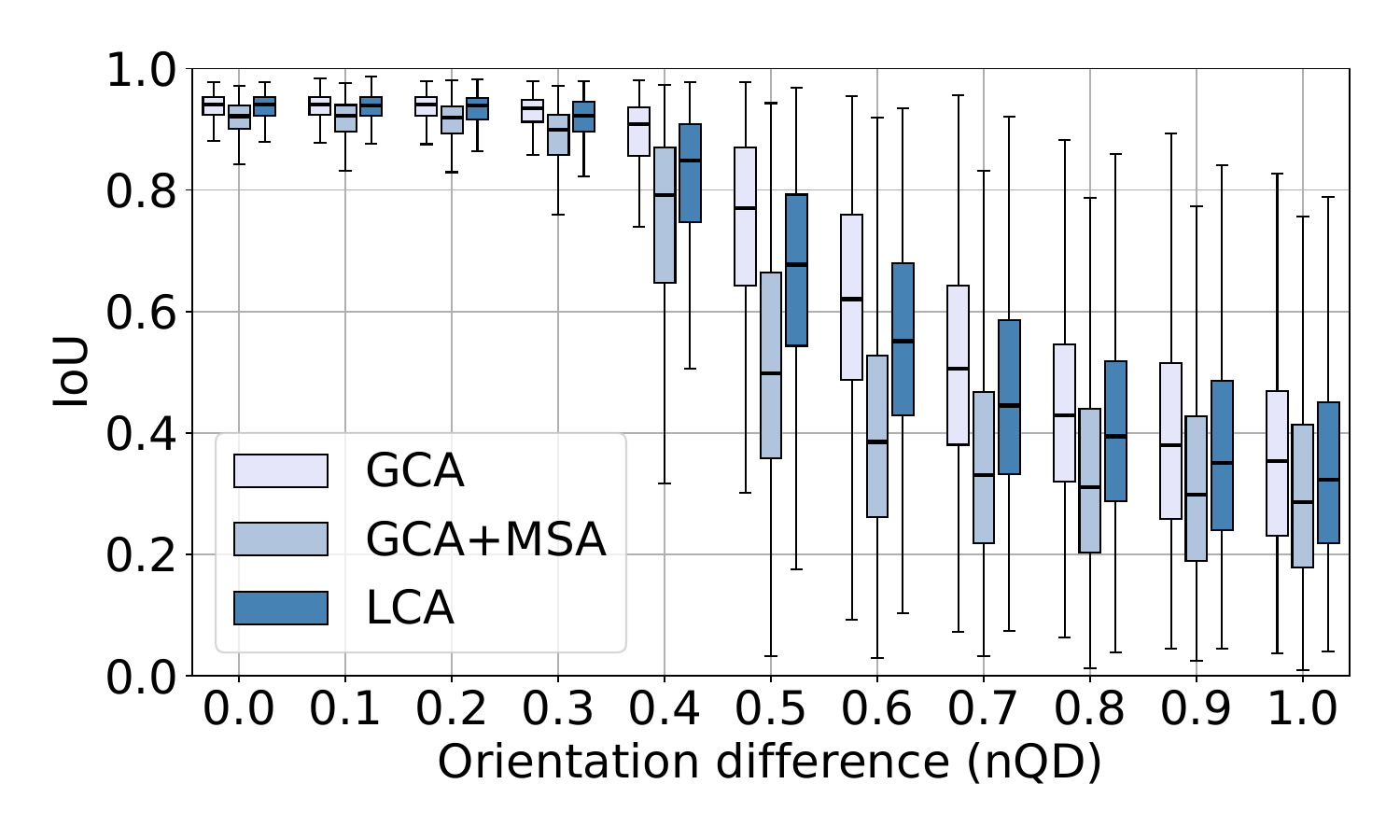}
    \caption{Same poses as training set, 7-10 parts difference between anchor and sample.}
    \label{fig:v2_quant_results_10part}
\end{subfigure}%
\begin{subfigure}{.5\textwidth}
  \centering
    \includegraphics[width = \linewidth]{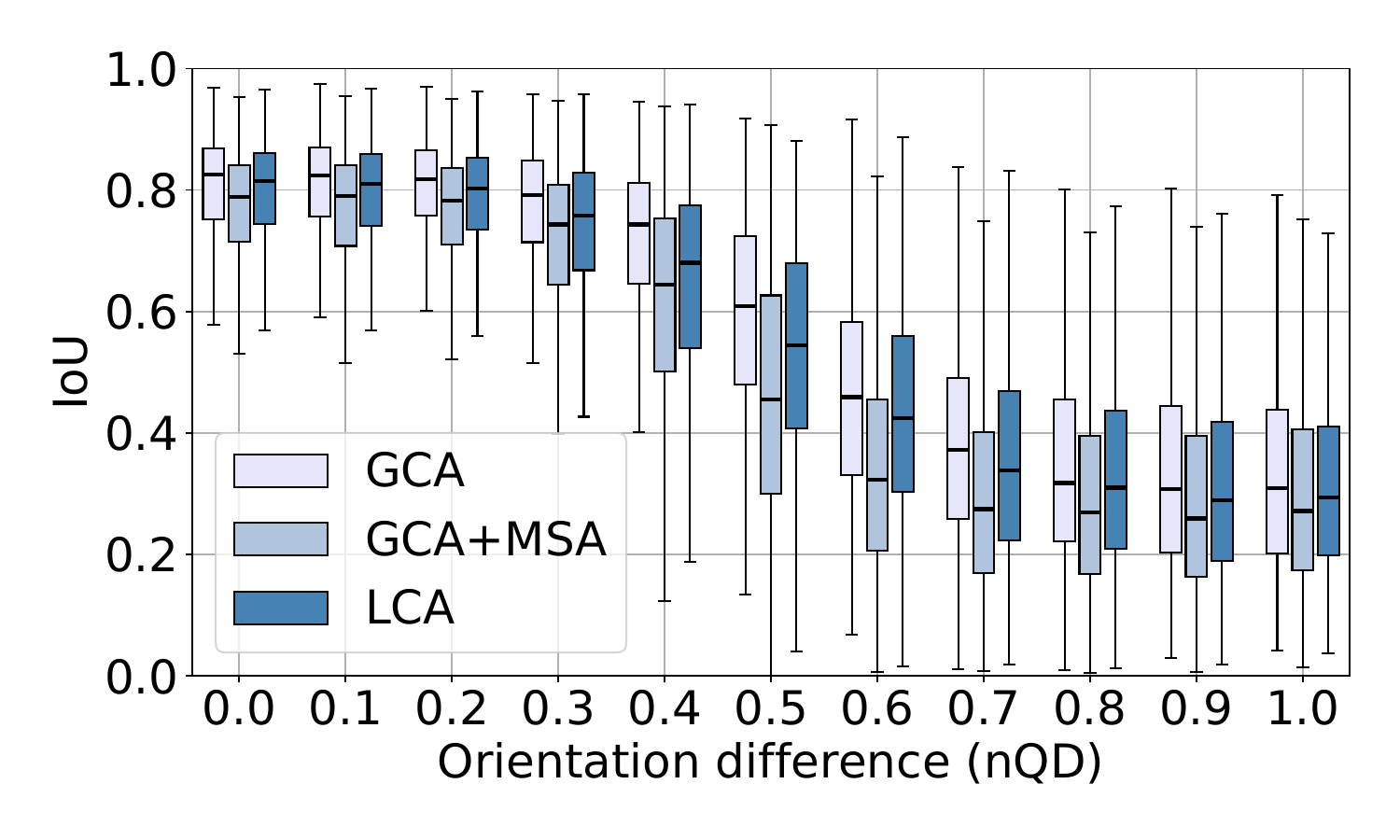}
    \caption{Different poses than training set, 7-10 parts difference between anchor and sample.}
    \label{fig:v2_extra_10part}
\end{subfigure}
\caption{Performance of three models with varying orientation difference, within the range of poses seen during training (a,c) as well as outside of this range (b,d), for various amounts of part difference between the anchor and sample image. When more parts differ between the images (c,d), the model scores a higher change IoU.}
\label{fig:supp_extra_part_diffs}
\end{figure}

\autoref{fig:test} contains our three models' performance on two test sets (the first containing the same poses seen during training, the second containing poses outside the range of poses seen during training). The reported IoU scores were given for image pairs containing 4~to~6 part-level differences that constitute change. For additional insight, \autoref{fig:supp_extra_part_diffs} contains the same plots when the amount of changed parts is less (1 to 3) or more (7 to 10). The same conclusions can be drawn as from \autoref{fig:test} regarding the comparison between the models at varying orientation differences. However, it can be seen that the IoU values are higher when there is more change present. We have observed that this score difference does not reflect a qualitative difference in the produced binary change masks when qualitatively inspecting test pairs. Thus, this higher IoU score for a greater number of parts constituting change is more likely due to a bias in the metric than a qualitative difference in the models' performance: for the same segmentation quality, image pairs that incur more change have a better IoU because the proportion of more difficult edge pixels to easier intra-part pixels decreases accordingly.

\subsubsection{Qualitative:}
\begin{figure}
    \centering
    \includegraphics[width=\textwidth]{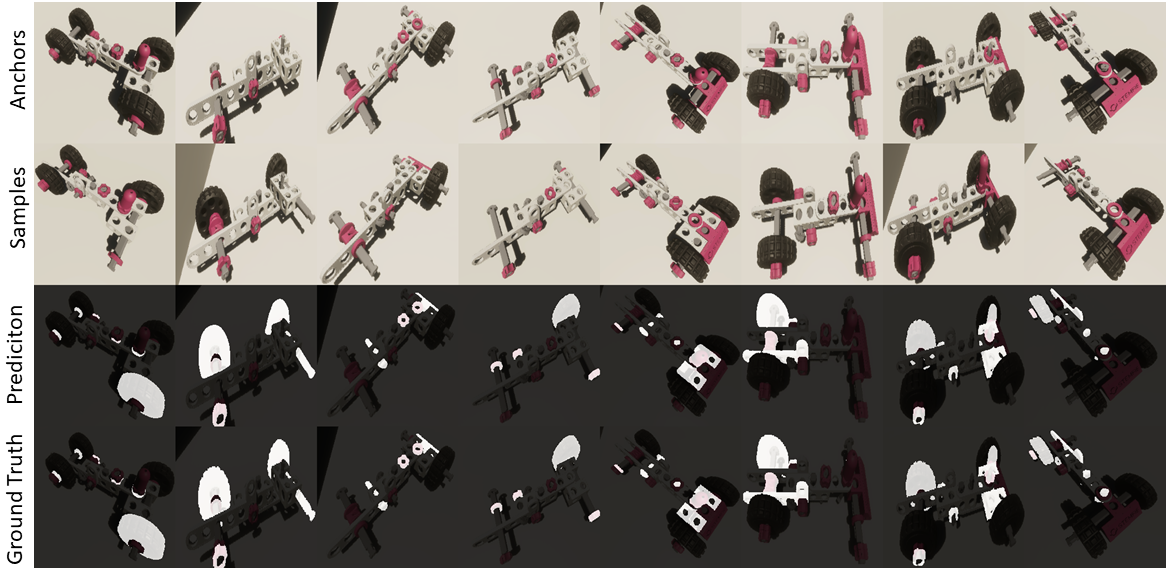}
    \caption{Qualitative results of some of the results presented in the GCA model on the test set with identical poses to the train set (corresponding to \autoref{fig:v2_quant_results}). The amount of change lies between 3 and 6 parts, and the orientation difference between pairs lies in the 0 to 0.1 nQD range.}
    \label{fig:Qual_results_v2}
\end{figure}

\begin{figure}
    \centering
    \includegraphics[width=\textwidth]{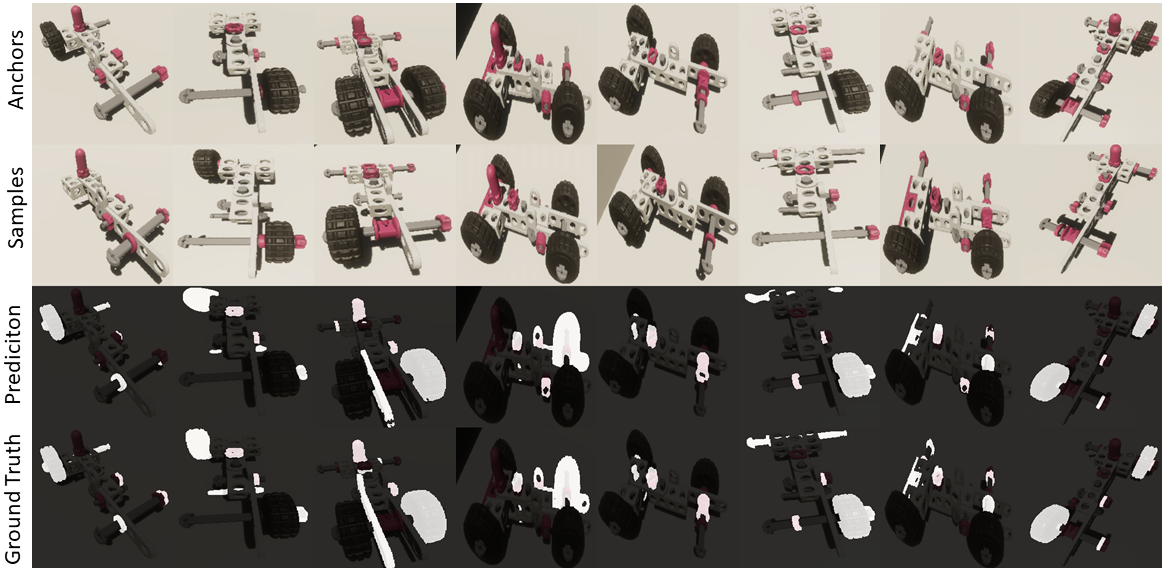}
    \caption{Qualitative results of the GCA model on test set with novel poses outside the range of poses in train set (corresponding to \autoref{fig:v2_extra}. The amount of change lies between 3 and 6 parts, and the orientation difference between pairs lies in the 0 to 0.1 QD range.}
    \label{fig:Qual_results_v2_extra}
\end{figure}

We additionally present here some qualitative results for the experiment presented in \autoref{fig:test} to provide a visualization of the performance drop between the poses present in the training set (see \autoref{fig:Qual_results_v2}) and the poses outside the range seen during training (see \autoref{fig:Qual_results_v2_extra}). By comparing the predicted change mask to the ground truth change mask, it can be seen that the GCA model exhibits near-perfect performance on images whose poses match the poses in the training set. However, on images whose poses fall entirely outside the range of poses in the training set (\autoref{fig:Qual_results_v2_extra}), a qualitative difference between predicted and ground truth change masks can be observed on some pairs. Nevertheless, the model still exhibits near-perfect performance on many pairs and manages to capture most of the relevant changes, albeit with poorer segmentation quality and more false positive and false negative predictions.

\FloatBarrier

\section{Qualitative results on ablation study}
\label{sec:supp_synth_qualitative}

We present here some qualitative results from our ablation study to show that, contrary to our GCA model, a concatenation-based (attention-less) model trained on perfectly aligned images manages to capture changes on parts that were never encountered as change during training.

\begin{figure}
    \centering
    \includegraphics[width=\linewidth]{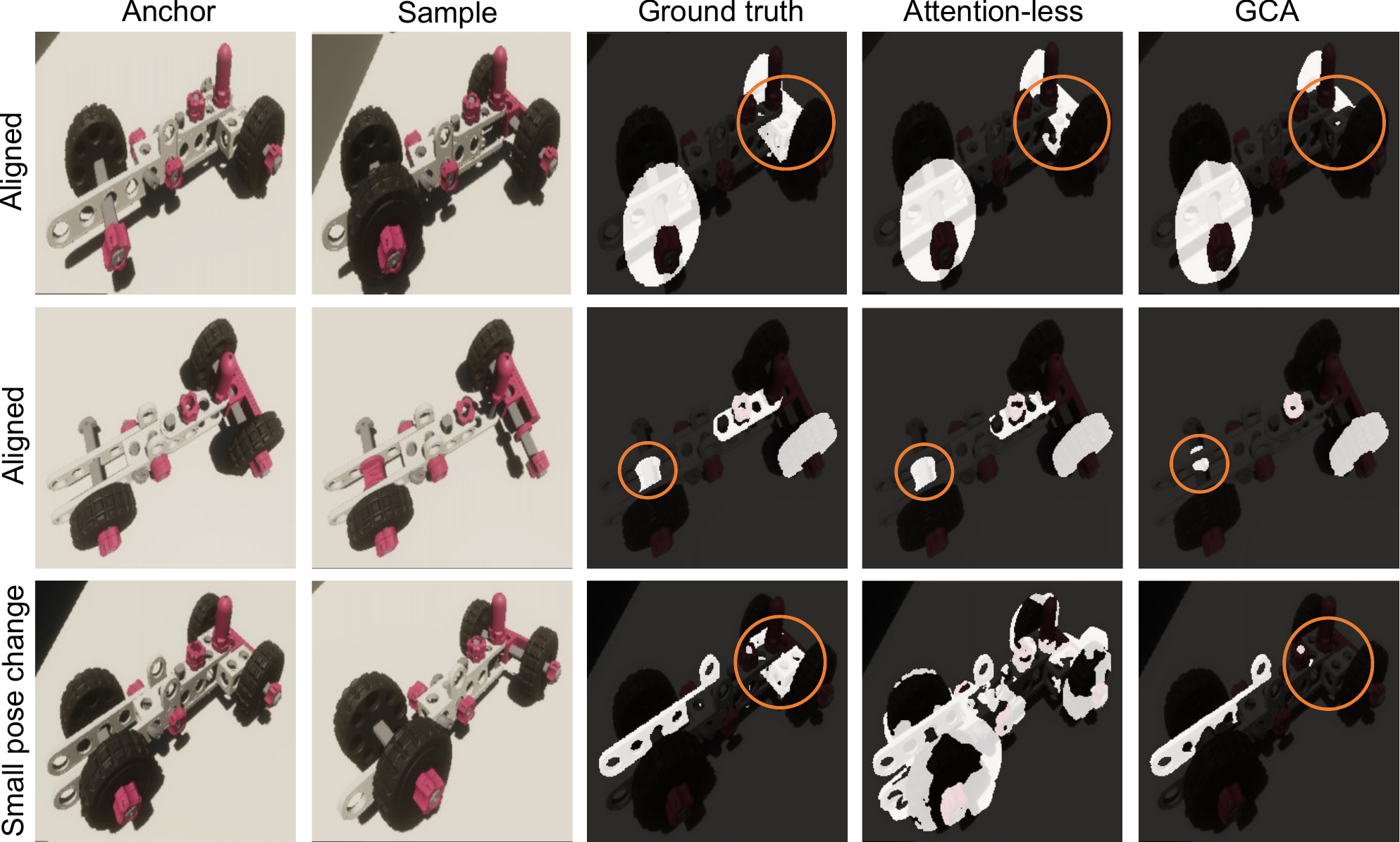}
    \caption{The attention-less model~(fourth column) refers to the concatenation-based model trained on perfectly aligned images, and GCA~(last column) refers to the cross-attention based model trained on small orientation differences. The orange circles highlight parts that never constitute change in the training set. The attention-less model is seemingly much more capable to capture change on those parts than the GCA-based model, which particularly fails to detect any change on the front bracket (rows 1 and 3). Note that the first two rows contain perfectly aligned image pairs and the last row contains an image pair with orientation difference. In the latter case, it can be seen that the attention-less (concatenation-based) model predicts a lot of change because it is only able to compare features across images at the exact same pixel coordinates, while the GCA model is able to capture meaningful change on the parts that it encounters during training.}
    \label{fig:qualitative_v1}
\end{figure}

\FloatBarrier 

\section{Additional qualitative results on real images}

\begin{figure}
    \centering
    \includegraphics[width=\linewidth]{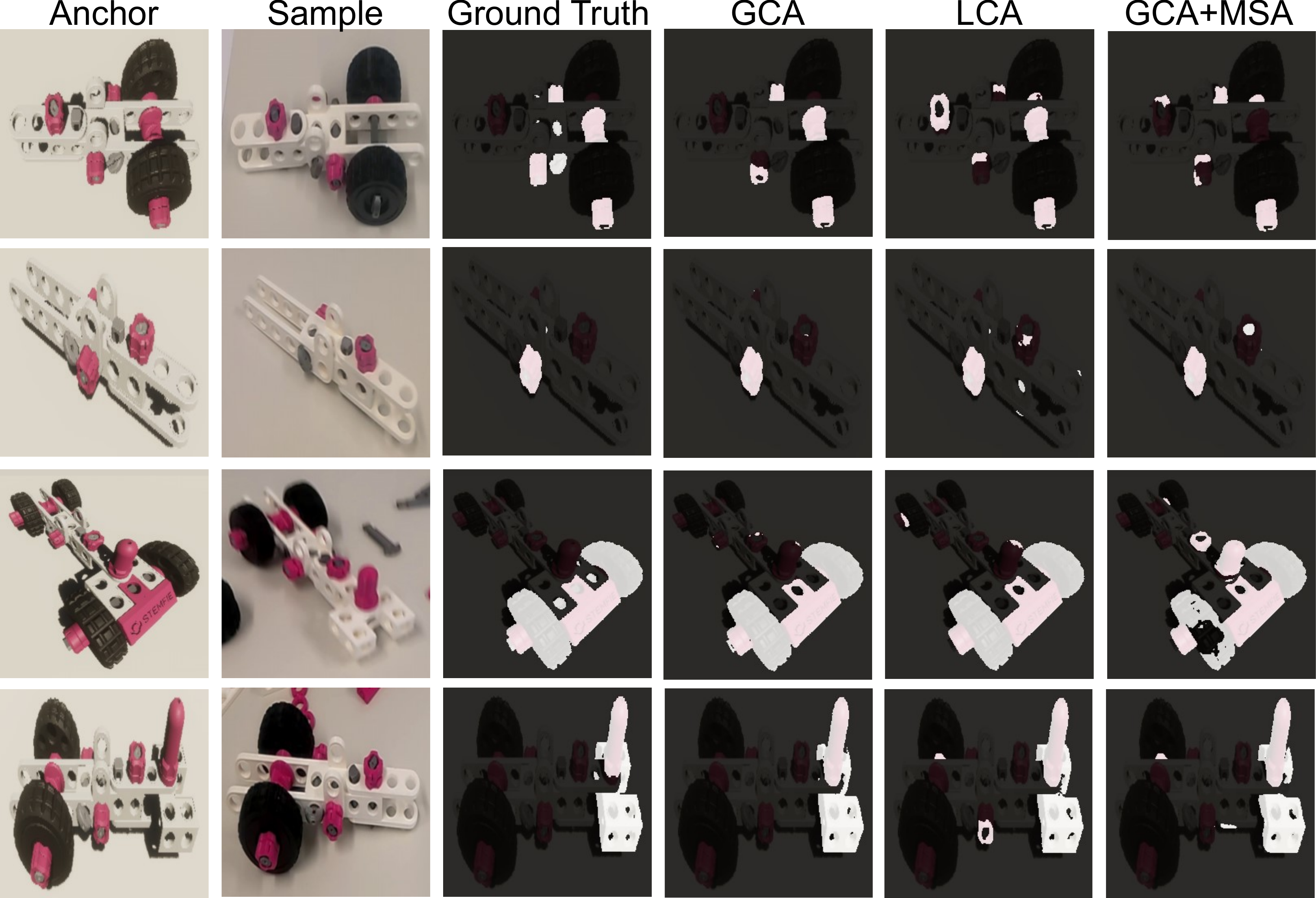}
    \caption{Additional qualitative results that further illustrate that the GCA model provides the highest quality predictions with the least amount of `noise' (false positives and false negatives). Nevertheless, the GCA model sometimes only notices parts of an error, such as in the first row where the misplaced nuts and washers are segmented, but not the misoriented pins themselves. Additionally, or in row 3, none of the models segment the long pin (visible through the holes in the front bracket).}
    \label{fig:additional_real}
\end{figure}

Finally, our paper only presents a handful of examples where our model segments errors from real-world images due to space constraints. We have added \autoref{fig:additional_real} here for additional examples and further analysis.

\end{document}